\documentclass{article}

\usepackage{microtype}
\usepackage{graphicx}
\usepackage{subfigure}
\usepackage{booktabs} %

\usepackage{hyperref}

\usepackage[ruled,vlined]{algorithm2e}
\usepackage{amsmath, amssymb}

\usepackage[accepted]{icml2024}

\usepackage{amsmath}
\usepackage{amssymb}
\usepackage{mathtools}
\usepackage{amsthm}
 
\usepackage[capitalize,noabbrev]{cleveref}

\theoremstyle{plain}

\theoremstyle{definition}

\theoremstyle{remark}

\usepackage[textsize=tiny]{todonotes}

\usepackage{xspace}    %
\usepackage{amsmath}   %
\usepackage{amsfonts}  %
\usepackage{amssymb}   %

\usepackage{graphicx}
\usepackage{subcaption}

\usepackage{booktabs}
\usepackage{multirow}
\usepackage{array}

\usepackage{enumitem}

\usepackage{ulem}

\usepackage{stfloats}  %

\usepackage{listings}
\usepackage{color}  
\usepackage{listings-verus} %

\usepackage{xcolor}

\lstdefinelanguage{Rust}{
  keywords={
    self,mut,fn,let,pub,impl,struct,enum,trait,for,if,else,while,loop,return,
    match,const,static,break,continue,unsafe,where,use,mod,as,in,type,move,
    box,dyn,await,async,try,true,false,
  },
  keywordstyle=\color{blue}\bfseries,
  sensitive=true,
  commentstyle=\color{gray}\itshape,
  morecomment=[l]{//},
  morecomment=[s]{/*}{*/},
  stringstyle=\color{purple},
  morestring=[b]",
  morestring=[b]',
  identifierstyle=\color{black},
  moredelim=[l][\color{green}]{//},
  moredelim=[s][\color{green}]{/*}{*/},
}

\usepackage{tcolorbox}

\usepackage{pgfplots}

\usepackage{tikz}
\usetikzlibrary{shapes.geometric, arrows, positioning}

\tikzstyle{arrow} = [thick,->,>=stealth]
\tikzstyle{process} = [rectangle, rounded corners, minimum width=3cm, minimum height=1cm,text centered, draw=black, fill=blue!30]
\tikzstyle{decision} = [diamond, minimum width=3cm, minimum height=1cm, text centered, draw=black, fill=red!30]
\tikzstyle{io} = [trapezium, trapezium left angle=70, trapezium right angle=110, minimum width=3cm, minimum height=1cm, text centered, draw=black, fill=green!30]
\tikzstyle{data} = [rectangle, minimum width=3cm, minimum height=1cm, text centered, draw=black, fill=yellow!30]

\definecolor{CBF1}{RGB}{255,99,132}  %
\definecolor{CBF2}{RGB}{54,162,235}  %
\definecolor{CBF3}{RGB}{255,206,86}  %
\definecolor{CBF4}{RGB}{75,192,192}  %
\definecolor{CBF5}{RGB}{153,102,255} %
\definecolor{CBF1b}{RGB}{205,89,112}  %
\definecolor{CBF2b}{RGB}{44,142,215}  %
\definecolor{CBF5b}{RGB}{133,92,225}  %

\newcommand{\ours}{\texttt{AlphaVerus}\xspace}

\definecolor{colorSean}{RGB}{200,0,0}       %
\definecolor{colorPranjal}{RGB}{72,61,139}  %
\definecolor{colorBryan}{RGB}{255,165,0}    %

\newcommand{\pranjal}[1]{}

\newcommand{\model}[1]{\texttt{#1}\xspace}

\newcommand{\llamal}{\model{LLaMA-3.1-70B}}
\newcommand{\llamas}{\model{LLaMA-3.1-8B}}

\newcommand{\gptl}{\model{GPT-4o}} 
\newcommand{\qwen}{\model{Qwen-32B}}

\newcommand{\dataset}[1]{\texttt{#1}\xspace}

\newcommand{\dafnybench}{\dataset{DafnyBench}\xspace}

\newcommand{\humaneval}{\dataset{HumanEval}\xspace}
\newcommand{\mbpp}{\dataset{MBPP}\xspace}

\icmltitlerunning{\ours{}}

\begin{document}

\twocolumn[
\icmltitle{\ours{}: Bootstrapping Formally Verified Code Generation through Self-Improving Translation and Treefinement}

\icmlsetsymbol{equal}{*}

\begin{icmlauthorlist}
\icmlauthor{Pranjal Aggarwal}{comp}
\icmlauthor{Bryan Parno}{comp}
\icmlauthor{Sean Welleck}{comp}
\end{icmlauthorlist}

\icmlaffiliation{comp}{Carnegie Mellon University}

\icmlcorrespondingauthor{Pranjal Aggarwal}{pranjala@cs.cmu.edu}
\icmlcorrespondingauthor{Sean Welleck}{wellecks@cmu.edu}

\icmlkeywords{Machine Learning, ICML}

\vskip 0.3in
]

\printAffiliationsAndNotice{}  %

\begin{abstract}
    
Automated code generation with large language models has gained significant traction, but there remains no guarantee on the correctness of generated code. 
We aim to use formal verification to provide mathematical guarantees that the generated code is correct.
However, generating formally verified code with LLMs is hindered by the scarcity of training data and the complexity of formal proofs. 
To tackle this challenge, we introduce \ours{}, a self-improving framework that bootstraps formally verified code generation by iteratively translating programs from a higher-resource language and leveraging feedback from a verifier. \ours{} operates in three phases: exploration of candidate translations, Treefinement---a novel tree search algorithm for program refinement using verifier feedback, and filtering misaligned specifications and programs to prevent reward hacking. Through this iterative process, \ours{} enables a \llamal{} model to generate verified code without human intervention or model finetuning.
\ours{} shows an ability to generate formally verified solutions for HumanEval and MBPP,
laying the groundwork for truly trustworthy
code-generation agents.\footnote{Code is available at \href{https://alphaverus.github.io}{https://alphaverus.github.io}}

\end{abstract}

\section{Introduction}

There has been an enormous effort to train code-generating large language models (LLMs)~\cite{human-eval,austin2021program,li2023starcoder,roziere2024codellamaopenfoundation,qwen2.5}, leading to LLM-powered agents that can perform tasks ranging from fixing bugs in software repositories to solving Olympiad-level algorithmic problems~\cite{Jimenez2023SWEbenchCL, Li2022CompetitionlevelCG}. 
Despite these successes, multiple studies have identified disturbing mistakes in LLM-produced code, including subtle bugs and serious security vulnerabilities~\cite{hendler2023understanding,Pearce2021AsleepAT,Jesse2023LargeLM,Zhong2023CanCR,perry2023insecure,Elgedawy2024OcassionallySA}.
Ultimately these mistakes stem from a fundamental property of LLMs: language models can generate any string of code, without regard to correctness.
As a result, automatically checking the correctness of LLM-generated code is one of the grand challenges facing the research community.

The generated code must be correct for all possible inputs it may receive. However,  today's code generation methods select or filter generations with imperfect proxies of correctness, such as runtime testing or human inspection.
Achieving perfect test coverage is typically infeasible~\cite{alphacode,liu2023is}, and incomplete coverage leads to an unreliable signal that can be exploited by a model~\cite{pan2022the,liu2023is,denison2024sycophancysubterfugeinvestigatingrewardtampering}. 
Relying on human review is equally problematic since it scales poorly and humans can struggle to tell whether LLM-generated code is correct~\cite{perry2023insecure}.
In turn, the difficulty of trusting generated code reduces the potential productivity gains from using LLMs and can lead to unexpected vulnerabilities or unreliable signals for improving models.

In contrast, generating code in a \textit{verification-aware programming language} such as Dafny~\cite{leino2010dafny}, F$^*$~\cite{fstar}, or Verus~\cite{10.1145/3586037} offers a promising approach to addressing these challenges by providing mathematical guarantees that a program obeys a specification for all possible inputs.
In this paradigm, code is paired with a specification and proof written in a specialized language, and a mechanical verifier checks whether the code meets the specification.
Doing so could dramatically improve the trustworthiness of the generated code: if the verifier passes, the LLM's generated program is mathematically guaranteed to meet the specification.
However, writing formal specifications and proofs introduces additional layers of complexity.
Furthermore, although LLMs have demonstrated success in automated theorem proving in mathematical domains~\cite{lu-etal-2023-survey,li2024a}, their capability to generate verified code for even basic algorithms is limited~\cite{clover,lohn2024minicodepropsminimalbenchmarkproving}. 

A significant barrier to automatically generating real-world, formally verified code is the scarcity of training data. 
In particular, verification-aware research languages have a rich history (e.g., Dafny~\cite{leino2010dafny}, F$^*$~\cite{fstar}), yet verifying real-world code in mainstream languages remains nascent.
For example, Verus~\cite{10.1145/3586037}--a verification language for the very popular language Rust--has fewer than 10 public repositories, despite Rust itself having millions of code examples. 
Hence, enabling formally verified code generation in a mainstream language such as Rust faces a \textit{bootstrapping problem}--how do we create an initial model that can generate even relatively simple verified programs, given the absence of training data?

\begin{figure*}[t]
    \centering
    \includegraphics[width=\linewidth]{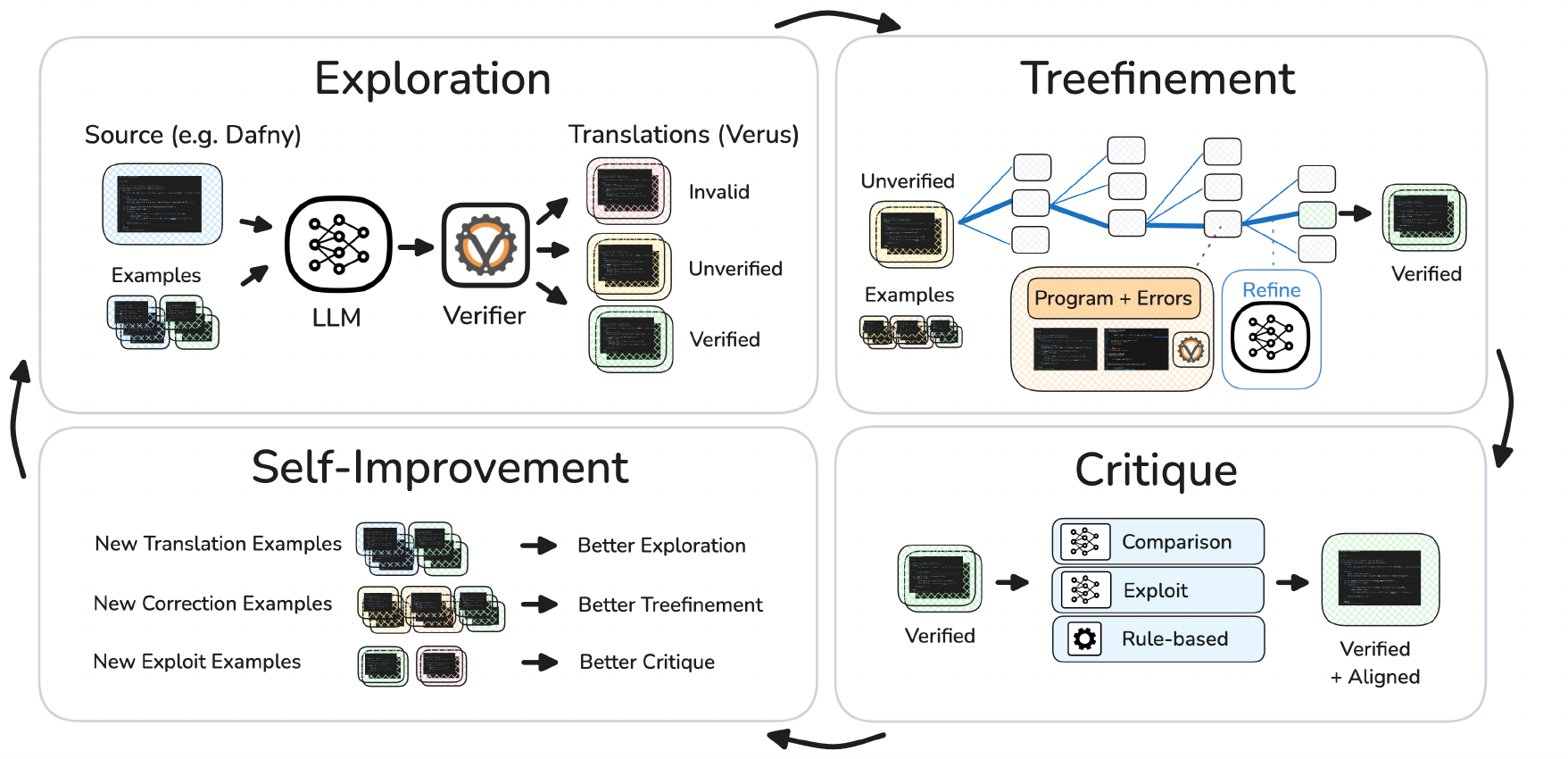}
    \caption{Overview of \ours{}, a self-improving framework for generating formally verified code. Each iteration consists of three key steps: (1) \textit{Exploration} translates programs from a source language to Verus by sampling multiple trajectories and selecting partially correct ones using verifier feedback, (2) \textit{Treefinement} iteratively fixes errors guided by verifier feedback and tree search, and (3) \textit{Critique} validates and filters out underspecified or incorrect translations. The framework bootstraps new exemplars after each iteration to continuously improve performance without human intervention. 
    }
    \label{fig:teaser}
\end{figure*}

We propose \ours{}, a framework for bootstrapping a formally verified code generation model by iteratively translating programs from a resource-rich domain and self-improving using feedback from the verifier. 
As illustrated in Figure~\ref{fig:teaser}, each iteration of \ours{} has three phases.
First, the \textit{exploration} phase generates candidate programs by translating from a source language (such as Dafny) to the target language (here, Verus) by generating multiple candidates and saving partially and completely verified attempts.
Second, \textit{Treefinement} refines the imperfect candidates through a novel tree search over the space of output programs using feedback from the verifier, saving the final verified program, along with its ancestors to serve as error correction examples.
We show that Treefinement leads to substantial gains over vanilla refinement strategies that resemble those used in concurrent work~\cite{yang2024autoverusautomatedproofgeneration,chen2024automatedproofgenerationrust}.
Third, \textit{critique models} detect misaligned translations and specifications--the one part of the pipeline that lacks formal guarantees.
Crucially, this alleviates \textit{reward hacking}, in which models learn to game the system by generating trivial or incomplete specifications, or even by identifying verifier limitations that cause trivial programs to pass the verifier. 
While previous work has investigated methods that rely on test cases~\cite{clover}, our critique models address the challenging problems of automated specification generation and validation without relying on any unit test cases. 

Each iteration of \ours{} collects new exemplars that improve the models in each phase, creating a cycle of improvement.
Thus, unlike recent work that relies on human experts to write correction prompts~\cite{yang2024autoverusautomatedproofgeneration}, our method requires no human intervention and automatically learns to generate better code. 
Moreover, the system operates using a single language model (e.g., Llama 70b), without the need for the expensive GPT-4 initialization used in concurrent work~\cite{chen2024automatedproofgenerationrust}.
Finally, the collected exemplars can be used to improve the verified code generation performance of any model without any finetuning. 

To demonstrate \ours{}, we consider Dafny~\cite{leino2010dafny} programs as the source domain, since the Dafny language has been around for over a decade and has accumulated a reasonable amount of code.
We run \ours{} to automatically collect the \textsc{Dafny2Verus-Collection}, a dataset of trajectories containing translated programs, error corrections, and critique examples based on the source dataset DafnyBench~\cite{loughridge2024dafnybenchbenchmarkformalsoftware}--a benchmark of 562 
programs of varying difficulty.
Finally, we evaluate the \ours{} pipeline by using the resulting data as few-shot exemplars for the downstream task of \textit{formally verified code generation}: generating complete, formally verified implementations—including both algorithmic code and proof annotations—given human-written specifications from independently developed benchmarks. 
Formally verified code generation is a significant step over concurrent work that focuses solely on the simplified, artificial setting of generating proof annotations for correct pre-written code~\cite{yang2024autoverusautomatedproofgeneration,chen2024automatedproofgenerationrust}. 
We show that \ours{} enables a Llama-70B model to successfully generate verified solutions to 33\% of Verified-HumanEval~\cite{human_eval_verus}, outperforming GPT-4o-based methods.
Furthermore, through ablations, we establish the necessity of each component in \ours{}.

In summary, our contributions are five-fold: (1) We propose \ours{}, a novel self-improving framework for generating formally verified code; (2) We present a novel combination of tree search and refinement that improves over time; (3) We propose a novel critique phase, which has to our knowledge the first neural method that can improve the quality of specifications without test cases; 
(4) We introduce a new dataset containing formally verified Verus programs, along with error pairs;  and (5) We demonstrate the effectiveness of our approach, evaluating its formally verified code generation abilities and ablating its components. 
In particular, \ours{} is the first method to achieve non-zero formally verified code generation performance on a verified version of HumanEval~\cite{human-eval}, thus establishing a starting point for code generation models that generate increasingly complex---yet trustworthy---code.

\begin{figure*}[t!]
\includegraphics[width=\linewidth]{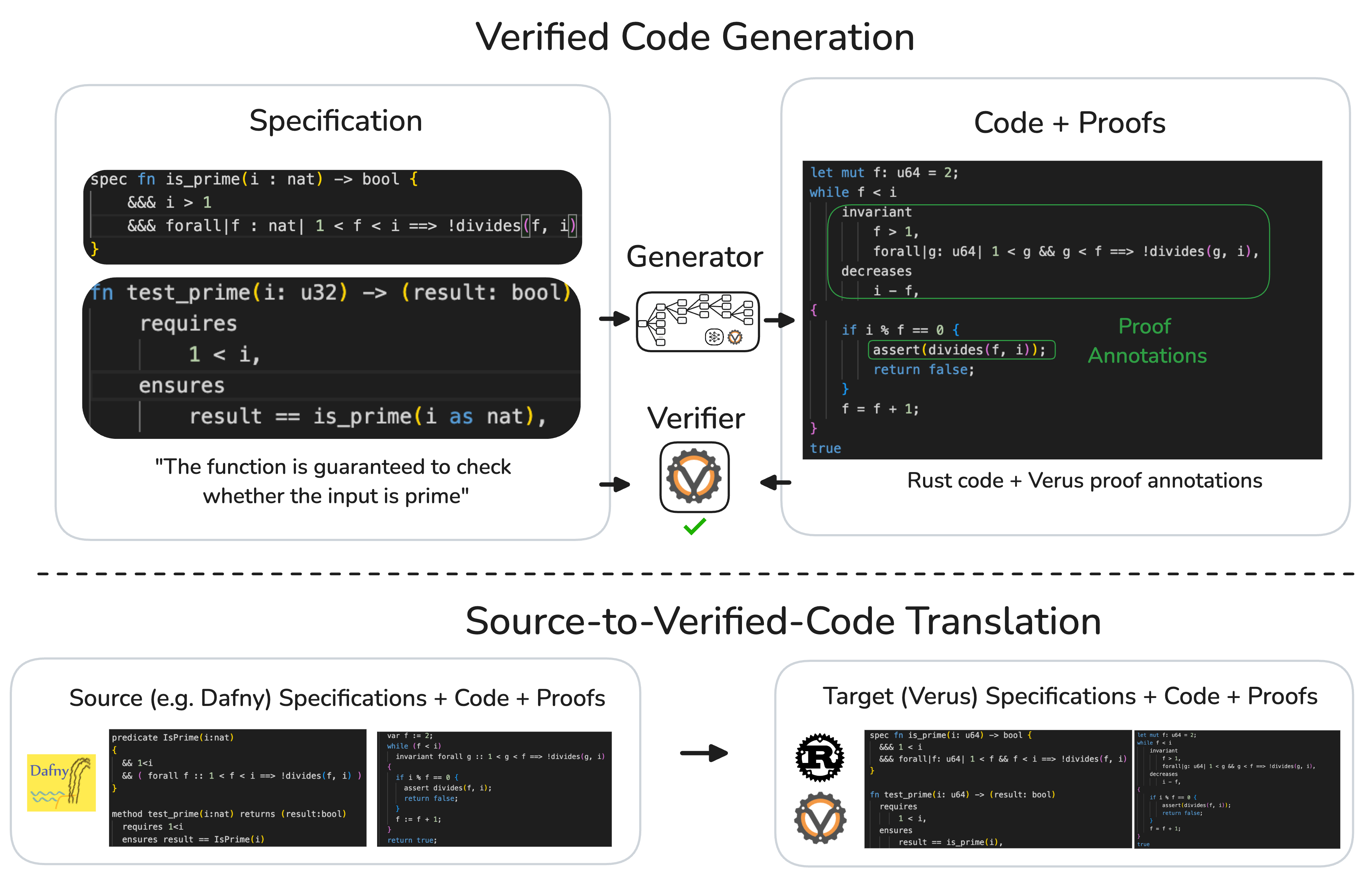}
    \caption{\textbf{Example of formally verified code generation and translation.} The figure shows the three key components of formally verified code: specifications (left), implementation with proof annotations (right), and the verifier. The specifications define a mathematical predicate \texttt{divides} and use it to specify primality. The implementation includes both algorithmic code and proof annotations (highlighted in green), which together allow the verifier to prove that the implementation satisfies the specification (or provides error messages upon failure).
    Translation consists of translating a source input (e.g., a Dafny program) into specifications, an implementation, and proof annotations.
    \ours{} performs each task using a generator consisting of exploration followed by Treefinement, and a subsequent critique step for translation.
    }
    \label{fig:verus-prime}
\end{figure*}

\section{Formally Verified Code Generation}

Our goal is to develop a model that generates formally verified code in a real-world programming language, which we refer to as \textit{formally verified code generation}.
Next, we provide background and then introduce \ours{}.

\begin{figure}
    \centering
    \includegraphics[width=0.99\linewidth]{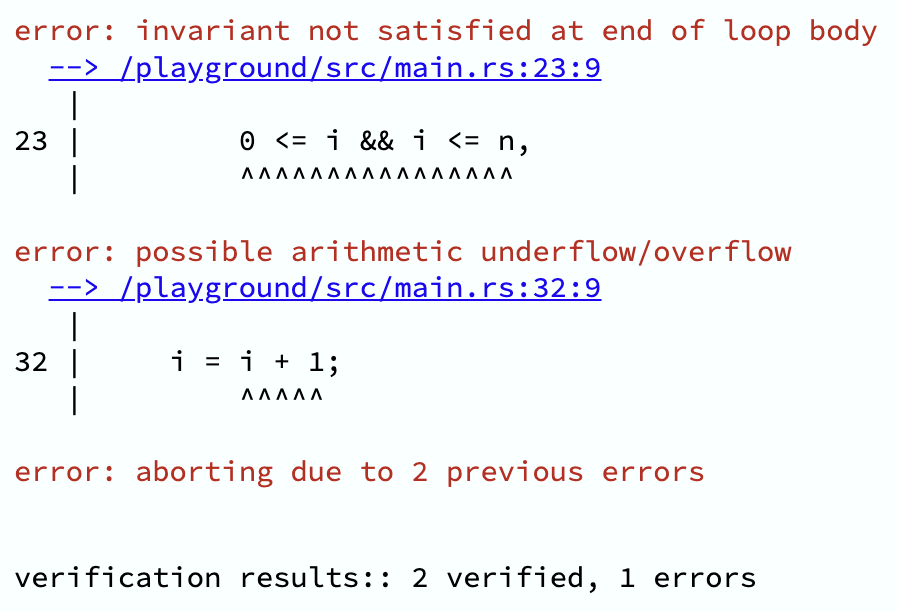}
    \caption{Example of Verus verifier errors showing two issues: (1) an invariant violation at the end of a loop body and (2) a potential arithmetic overflow in an increment operation. Errors point to exact lines and the verification results indicate 2 successful and 1 failed verifications. %
    }
    \label{lst:errors}
\end{figure}

\paragraph{Formal verification of code.}
Formal verification ensures that a program adheres to a formally defined specification of its intended behavior. As illustrated in Figure~\ref{fig:verus-prime}, formally verified code typically consists of three components: (1) formal specifications $y_S$ defining the expected input-output behavior; (2) a code implementation $y_I$ intended to satisfy the specifications; and (3) a proof $y_P$ demonstrating that the implementation conforms to the specifications. A verifier $v(y_S,y_I,y_P)\rightarrow \{0,1\}$ uses the proof to statically check that the implementation meets the specification for all possible inputs, returning 1 when the program is correct with respect to $y_s$ and 0 when verification fails.
Upon failure, the verifier additionally returns a set of messages $\{m_1,\ldots,m_M\}$ containing the number of verified statements, the number of errors, and localized error messages (e.g., see~\autoref{lst:errors}).

\paragraph{Misaligned specs and implementations.} The specifications themselves are not verified, as they represent the developer's intended behavior. Therefore, it is critical that the specifications accurately reflect the desired input-output behavior for all possible inputs.
We use the term \textit{misaligned} to refer to situations in which the specification does not reflect the desired input-output behavior.
This includes (i) a misalignment between the specification and the developer's intent, such as missing an edge case or allowing a trivial implementation, and (ii) a misalignment between the specification and an implementation, when the implementation passes the verifier but does not implement the functionality in the specification.
The latter can occur due to language features in verification-aware languages that cause programs to pass the verifier (such as writing ``\texttt{assume (false)}'', which causes any program to pass). 

\paragraph{Formally verified code generation.}
Our goal is to develop a model that generates formally verified code given a specification.
Specifically, 
\begin{align}
    (y_I,y_P)\sim G\left(y_S;c, \theta\right),
\end{align}
where $G(\cdot)$ is a generation algorithm such as sampling from a language model with parameters $\theta$, and the model generates both an implementation $y_I$ and proofs $y_P$ given a specification $y_S$ and any additional context $c$.
The goal is for the resulting code to verify, i.e., $v(y_S, y_I, y_P) = 1$. 

\paragraph{Bootstrapping formally verified code generation.}
A practical goal is to perform formally verified code generation in a mainstream language, such as Rust code verified with the Verus verifier~\cite{10.1145/3586037}.
However, doing so raises a technical challenge: it is infeasible to train a model on $(y_S,y_I,y_P)$ examples since such examples do not exist.
We refer to this as a \textit{bootstrapping problem}, since we need to create an initial generation model (that we may subsequently improve) without any training data.
Next, we describe \ours{}, a framework for bootstrapping a verified code generation model by translating from a more resource-rich language.

\section{\ours{} for Bootstrapping Formally Verified Code Generation}

To enable verified code generation in the absence of training data in our target language (Verus), we propose to iteratively translate programs from a higher-resource domain into Verus.
Each iteration collects data by exploring candidate translations, refining them with a novel tree search, then filtering out misaligned programs. 
Finally, we use the data to enable a verified code generation model (via few-shot learning), and evaluate the model plus the tree search on the downstream task of verified code generation: generating verified code and proofs given a held-out test specification.

\subsection{Translation}

\ours{} translates programs using a three-stage pipeline consisting of \textit{exploration}, \textit{refinement}, and \textit{critique}.
The exploration stage translates source programs into candidate Verus programs. The refinement stage repairs the programs using a novel tree search over program refinements. The critique stage uses a suite of models to discard flawed specifications and implementations that could degrade future iterations.
The pipeline iterates, creating a self-reinforcing cycle where verified programs and refinement trajectories improve the models' capabilities, enabling translation of increasingly complex programs.
The result is a growing synthetic dataset of progressively more complex and reliable Verus programs.
The complete algorithm is listed in Algorithm~\ref{alg:translation} and visualized in Figure~\ref{fig:teaser}.

\paragraph{Exploration.}
Given a source program $x$ (e.g., a Dafny implementation, specification, and proofs), exploration uses a model to generate candidate target (i.e., Verus) programs:
\begin{align}
    \{y_1,\ldots,y_k\} & \sim G_{explore}\left(x;D^{(i)}_{x\rightarrow y}\right),
\end{align}
where $G$ is a generation algorithm (e.g., LLM sampling) that is given the source and a set of (source, target) examples $D^{(i)}_{x\rightarrow y}$.
Initially, $D^{(0)}_{x\rightarrow y}$ has a few hand-written examples.

Any generated (source, verified program) pairs are placed in a candidate set, $C$, that will be passed to the filtering stage. If no candidates verify for source $x$, candidates that are syntactically correct proceed to refinement.
Intuitively, this stage serves as initial ``exploration'', in that it generates a set of candidates that may eventually be refined and filtered into verified programs in the later stages. 
Unlike other methods of bootstrapping~\cite{zelikman2022starbootstrappingreasoningreasoning, lin2024leanstarlearninginterleavethinking} that discard anything but correct solutions, we use both syntactically correct programs and fully verified programs for further improvement, expanding the learning signal.

\paragraph{Refinement with Treefinement.}
Having a verifier opens the possibility of refining candidate programs into verified ones by providing detailed feedback, including unverified functions and specific errors like overflows, unsatisfied conditions, and syntactic mistakes (e.g.,~\autoref{lst:errors}). 
While human programmers often use such feedback for iterative corrections, naively providing LLMs with incorrect solutions and feedback often fails to produce improvements.
Our key insight is that verifier feedback induces an implicit ordering of solutions based on verified functions and error severity.
This ordering lets us extend common refinement techniques by framing refinement as a tree search over the space of refined programs, which we call 
\textit{Treefinement}.

Specifically, the refinement stage takes syntactically correct but unverified candidate translations $\{y_1,\ldots,y_{k'}\}$ and performs a tree search to discover verified programs. Each node in the tree contains an imperfect program and its associated errors, $(y, e(y))$. Nodes are expanded by invoking a refinement model:
\begin{align}
    \{y_1',\ldots,y_k'\} & \sim G_{refine}\left(y, e(y);D^{(i)}_{y\rightarrow y'}\right),
\end{align}
where $D^{(i)}_{y\rightarrow y'}$ is a set of (program, error, correct program) examples, initially containing a few hand-written examples.

Given a node scoring function $v(y)\rightarrow \mathbb{R}$ that is used to prioritize nodes, we can search over the space of program refinements with a tree search algorithm that selects and expands nodes, such as breadth-first or depth-first search.

We develop a symbolic scoring function based on the number of (un)verified functions, errors, and warnings:
\begin{align*}
s(y) = \frac{ n_{\text{ver}}( y ) - \alpha n_{\text{err}}( y ) - \beta n_{\text{warn}}( y ) }{ n_{\text{ver}}( y ) + n_{\text{unver}}( y ) }
\label{eq:score}
\end{align*}
where \( n_{\text{ver}}( y ) \) is the number of verified functions in \( y \), \( n_{\text{err}}( y ) \) and \( n_{\text{warn}}( y ) \) are the counts of errors and warnings from the verifier for the node's program \( y \). \( \alpha \) and \( \beta \) are hyperparameters controlling the penalties for errors and warnings, respectively. \autoref{lst:errors} shows example verifier feedback, with $n_{\text{ver}}(y)=2$, $n_{\text{unver}}(y)=1$, $n_{\text{err}}(y)=1$, and $n_{\text{warn}}(y)=0$, and its score is computed as $(2- \alpha \cdot 1 - \beta \cdot 0)/(2+1)$.
Intuitively, programs that are closer to a verified program have higher scores, with proximity determined by the proportion of verified functions, resolved errors, and resolved warnings.
Upon generating a verified program, the program's search trajectory is added to a candidate set $C_{\tau}$, and the new (source, program) pair to the candidate set $C$ that is passed to the critique stage.

Treefinement extends two kinds of prior methods into a new search over program refinements. 
First, refining LLM outputs is a common technique~\cite{madaan2023selfrefine, kamoi-etal-2024-llms}, but not within a tree search.
On the other hand, tree search developed in step-by-step mathematical problem solving involves appending solution steps rather than refining a full program~\cite{wu2024empiricalanalysiscomputeoptimalinference}. 
Our approach specifically addresses the non-local nature of error fixes.

Although Treefinement can use any tree search algorithm, we use REBASE (REward BAlanced SEarch)~\cite{wu2024inference}. 
REBASE allocates an exploration budget by sampling nodes from a distribution determined by the node scores at the current depth, providing an effective balance of exploration and exploitation. 
The search continues until it finds a verified program or reaches a maximum depth.

\begin{table*}[ht]
    \centering
    \small
    \begin{tabular}{p{0.18\textwidth}p{0.27\textwidth}p{0.27\textwidth}p{0.18\textwidth}}
    \toprule
    \textbf{Stage} & \textbf{Feedback} & \textbf{Model} & \textbf{Data Collected} \\
    \midrule
    Exploration & Verifier (errors) & LLM + Parallel Sampling& Verified Translations \\
    \midrule
    Treefinement & Verifier (value), Verifier (errors) & LLM + Tree Search  + Refinement& Error Fix Triplets, Verified Translations \\
    \midrule
    Critique Module & 
    Rules, Trivial Programs, Verifier (binary), Comparison LLM
    & 
    Regex, String Manipulation, Prompted LLM, Exploit LLM
    & 
    Exploit Pairs \\
    \bottomrule
    \end{tabular}
    \caption{Different components used in iterative translation in \ours}
    \label{tab:components}
\end{table*}

\paragraph{Critique.} Synthesized specifications are the one part of the translation pipeline that lacks formal guarantees,
which can result in a mismatch between the intended and actual functionality of generated programs.
Furthermore, in a few degenerate cases, there can be a mismatch between the specification's intent and the program's implementation, since Verus has features that can result in trivial programs passing the verifier (e.g., \texttt{assume (false)}).
These can lead to reward hacking, causing a snowballing effect when used as exemplars in future iterations.
Hence, we propose a three-part approach for filtering out such misaligned programs: a \textit{rule-based model}, a \textit{comparison model}, and a \textit{exploit model}.

The rule-based model receives a generated program $y$, and detects if $y$ uses a Verus feature which leads to a trivial program. 
Since there are a relatively small number of such features, and these features can be detected through string matching, it suffices to use a list of hand-coded filters. This includes checking for \texttt{assume (false)}, ``\texttt{\#[verifier::external]}'', and trivial preconditions.

The comparison model $f(x,y)$ receives a source input $x$ and a program candidate $y$, and evaluates
whether the specifications and algorithms used in the candidate match those
 from the source in intent and structure.
In practice, we prompt a model to generate multiple evaluation sequences and reject an output if at least $r$ sequences indicate rejection.

The exploit model is an adversarial approach that leverages the feedback from Verus. We use a generator prompted to generate simple and often trivial solutions--such as returning an empty array--that satisfy the specifications, i.e.,
\begin{align}
    (y_I,y_P) \sim G_{\text{exploit}}\left( y_S ; D^{(i)}_{\text{exploit}} \right),
\end{align}
where \( y_S \) is a generated specification, \( y_I,y_P\) is an implementation and proof, and \( D^{(i)}_{\text{exploit}} \) contains (specification, implementation+proofs) examples.
If such simple solutions pass verification, it indicates that the specification is flawed, and the corresponding translation is discarded. 
This includes subtle forms of mis-specification; for instance, on tasks requiring array manipulation, the synthesized specification may omit conditions on the array length, which results in trivial solutions such as returning an empty array.

\paragraph{Iteration.}
\label{sec:iteration}

Finally, the newly generated programs and a subset of the error trajectories are added to a pool of data that is used by the translator, refinement, and critique models in the next iteration.
In this sense, the models ``self-improve'' given access to the Verus environment, so long as the generated examples are useful exemplars.

Formally, for exploration, we create a new pool of examples,
\begin{align}
    D^{(i+1)}_{x\rightarrow y} =  D^{(i)}_{x\rightarrow y} \cup \tilde D^{(i+1)}_{x\rightarrow y},
\end{align}
where $\tilde D^{(i+1)}_{x\rightarrow y}$ consists of the (source, program) candidates $C$ that were collected during exploration and refinement, and that additionally pass the critique stage.

For refinement, we create a new pool of examples using the successful trajectories $C_\tau$ collected during refinement. 
Namely, we keep those trajectories whose final program passes the critique stage, and pair each intermediate program $y$ and its errors with the final program $y'$, i.e.,
\begin{align}
    \tilde D^{(i+1)}_{y\rightarrow y'} = \{(y, e(y), y') \mid y \text{ is an ancestor of } y'\},
\end{align}
and set $D^{(i+1)}_{y\rightarrow y'}= D^{(i)}_{y\rightarrow y'}\cup \tilde D^{(i+1)}_{y\rightarrow y'}$.

Similarly, for the exploit model, we add (specification, program) exploits that pass the verifier into $D^{(i+1)}_{\text{exploit}}$ to be used by the exploit model in the next iteration. Table~\ref{tab:components} summarizes the components, feedback sources, models, and generated synthetic data at each stage. \pranjal{Should I mention dafny2verus collection here?}

To use synthetic data as in-context exemplars, we employ a stochastic few-shot sampling approach. Specifically, each time a generator is called, we randomly sample $k$ examples from its respective data pool. This method reduces the computational cost associated with fine-tuning large models and, as shown in our results, enables other models to leverage the data pool to improve their performance without any training. Nevertheless, fine-tuning models and developing learning objectives remain interesting future directions.

\paragraph{Source domain: Dafny.}

As our initial source domain, we consider Dafny--a language that follows a similar paradigm to Verus and has been in use for over a decade, resulting in a larger set of available data. Specifically, we use the \dafnybench{} dataset, which contains 562 non-trivial programs suitable for translation.

Translating Dafny programs to Verus presents several challenges due to two major differences: 1. \textit{Language Constructs:} Significant differences exist in supported features, data types, and the design of the underlying verifier, rendering direct translations infeasible. 2. \textit{Proof Requirements:} Verus imposes more rigorous proof obligations, such as overflow checks, making proofs harder to verify.

\subsection{Downstream Evaluation}
\label{ssec:downstream}

After generating high-quality synthetic data in the form of formally verified Verus programs and error-feedback-correction triples, we use the data to enable a model that performs formally verified code generation.

We adopt a two-stage approach comprising \textit{exploration} and \textit{Treefinement}. In the exploration stage, given a specification $y_s$, we generate $k$ candidate programs $\{ y^{(1)}, \ldots, y^{(k)} \}$.
If any candidate passes verification, we consider the task solved. If none succeed, we initialize Treefinement with the candidates and run it until we obtain a fully verified solution or reach a maximum number of iterations.
Conceptually, this can be viewed as a meta-generator that uses the collected data as a source of few-shot exemplars,
\begin{align}
    (y_I,y_P)\sim G(y_S;D_{y},D_{y\rightarrow y'}),
\end{align}
which means generating an implementation and proofs using a language model prompted with a subset of the collected verified programs $D_y$ and a test specification $y_S$, followed by Treefinement with the collected refinement examples $D_{y\rightarrow y'}$.
In practice, we randomly select a subset of exemplars before each call to the generator.

This generator is then evaluated using the verification success on a benchmark of held-out specifications, i.e.
\begin{align}
\text{pass@}K(G,\{y_S^1\ldots y_S^{N}\})=\frac{1}{N}\sum_{i=1}^{N}v(y_S^i,G(y_S^i)), \label{eq:passk}
\end{align}
where $K$ is the total number of sequences generated, and here $v(\cdot,\cdot)$ returns 1 if any of the generated programs pass.

Unlike prior work that requires LLMs to fill proof annotations in existing code and specifications~\cite{loughridge2024dafnybenchbenchmarkformalsoftware, yang2024autoverusautomatedproofgeneration,chen2024automatedproofgenerationrust}, we evaluate our models on this challenging task of generating both the code and the corresponding proofs given only the specifications. 
This task is significantly more complex, as the LLM must structure the code to facilitate the completion of the proof.

\section{Experimental Setup}

\paragraph{Generators.} We use \llamal{} for translation experiments and additionally evaluate \llamas{}, \qwen{}, and \gptl{} for downstream tasks. 
The exploration phase uses $k = 256$ samples, while tree search uses breadth $32$ and maximum depth $8$. 
The exploit model generates $32$ responses per specification. All generators use nucleus sampling with temperature $0.7$.

\paragraph{Translation.} We use \dafnybench{} as our source domain $D_{src}$ for our translation experiments. Starting with 782 programs, we filter to 562 by excluding those that verify without proof annotations. The exploration model $G_{explore}$ is initialized using a Verus syntax file and 5 examples from the Verus repository. See Appendix~\ref{app:exp} for details.

\paragraph{Downstream Evaluation.} We evaluate on the task of formally verified code generation, where models must generate both an implementation and proof annotations given a specification. Performance is measured using Pass@K as defined in the \autoref{eq:passk}, where success requires at least one correct solution of the $K$ generated programs.

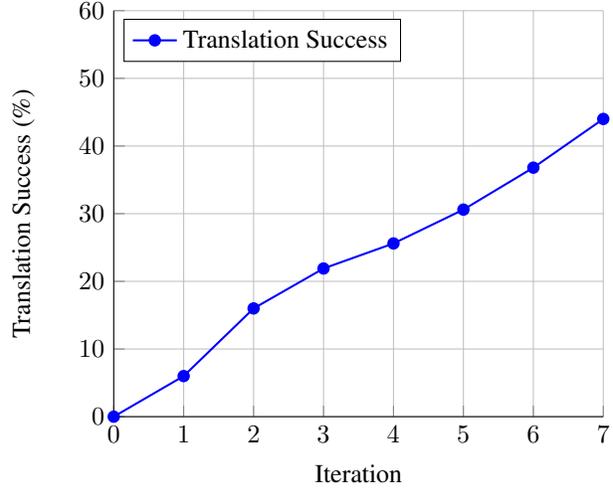
\begin{figure}[t]
    \centering
    \resizebox{\linewidth}{!}{%
    \begin{tikzpicture}
    \begin{axis}[
        width=\linewidth,
        xlabel={Iteration},
        ylabel={Translation Success (\%)},
        axis y line*=left,
        axis x line*=bottom,
        xmin=0, xmax=7,
        ymin=0, ymax=60,
        ytick={0,10,20, 30, 40, 50, 60},
        grid=major,
        legend style={
            at={(0.02,0.98)},
            anchor=north west,
            legend columns=-1,
            /tikz/every even column/.append style={column sep=0.5cm}
        },
        xtick={0,1,2,3,4,5,6,7},
        ]
        
        \addplot[blue, thick, mark=*] coordinates {
            (0,0)
            (1,6)
            (2,16)
            (3,21.9)
            (4,25.6)
            (5,30.6)
            (6,36.8)
            (7,44.0)
        };
        \addlegendentry{Translation Success}

    \end{axis}
    
    \end{tikzpicture}
    }
    \caption{\textbf{Programs Translated over Iterations}. The translation success rate shows consistent improvement over iterations.}
    \label{fig:translation}
\end{figure}

\paragraph{Datasets:} We evaluate on verified versions of \mbpp{} and \humaneval{} datasets. In particular, \mbpp{}-verified is sourced from \cite{yang2024autoverusautomatedproofgeneration, misu2024towards} and contains 78 programs from the original MBPP dataset~\cite{austin2021program}. \humaneval{}-Verus is sourced from a concurrent open-source effort~\cite{human_eval_verus} to translate existing \humaneval{} programs to Verus. Since each task in \humaneval{}-Verus is typically implemented and verified using multiple functions, we split each program into individual provable functions, ensuring that all dependent functions needed are present. Specifically, we split 49 programs into 85 functions and evaluate methods on these 85 functions. We use a snapshot from November 4th, 2024 with commit hash \texttt{ddb9ba3}. For brevity, we refer to \humaneval{} and \mbpp{} as their respective verified versions throughout this paper.

\paragraph{Baselines:} Our primary evaluation is performed on verified code generation. Since no existing baselines exist for the task, we use few-shot variants (Listing~\ref{lst:verus_inference}) of base models. We tried our best to adapt AutoVerus~\cite{yang2024autoverusautomatedproofgeneration} to verified code generation. However, due to the complexity of its hand-written prompts, we were not able to achieve non-trivial performance and therefore did not include it.
To compare with prior methods, we test \ours{} on the \mbpp{} proof annotation task against SAFE++~\cite{chen2024automatedproofgenerationrust} and AutoVerus~\cite{yang2024autoverusautomatedproofgeneration}.

\begin{table*}[t]
    \centering
    \begin{tabular}{lcc}
    \toprule
    \textbf{Method} & \textbf{HumanEval} & \textbf{MBPP} \\
    \midrule
    \multicolumn{3}{l}{\textit{Baselines}} \\
    \midrule
    GPT-4o & 27.1\% & 35.9\% \\
    Llama 3.1 70B & 11.8\% & 26.9\% \\
    \midrule
    \multicolumn{3}{l}{\textit{Ablations (Treefinement Variants)}} \\
    \midrule[\heavyrulewidth]
    Single-Turn Linear Self-Refine $b=256,d=1$ & 29.4\% & 61.5\% \\
    Multi-Turn Linear Self-Refine $b=32,d=8$ & 29.4\% & 62.8\% \\
    Best-First Search $b=32,d=8$ & 28.2\% & 61.5\% \\
    \midrule[\heavyrulewidth]
    \multicolumn{3}{l}{\textbf{\ours{}} (Llama 3.1 70B)} \\
    \midrule
    Exploration & 27.1\% & 59.1\% \\
    + Treefinement (Rebase, $b=32,d=8$)& \textbf{32.9\%} & \textbf{65.7\%} \\
    
    \bottomrule
    \end{tabular}%
    \caption{\textbf{Verified code generation performance} on the \humaneval{} and \mbpp{} benchmarks (pass@256). The highest accuracy is in bold. \ours{} and its variants use examples from \textsc{Dafny2Verus-Collection} to enable verified code generation (\S\ref{ssec:downstream}). \ours{} (i.e., exploration followed by Treefinement with Rebase) achieves the best performance. 
    }
    \label{tab:combined}
\end{table*}

\begin{figure*}[t]
    \centering
    \begin{minipage}[t]{0.49\textwidth}
    \includegraphics[width=\textwidth]{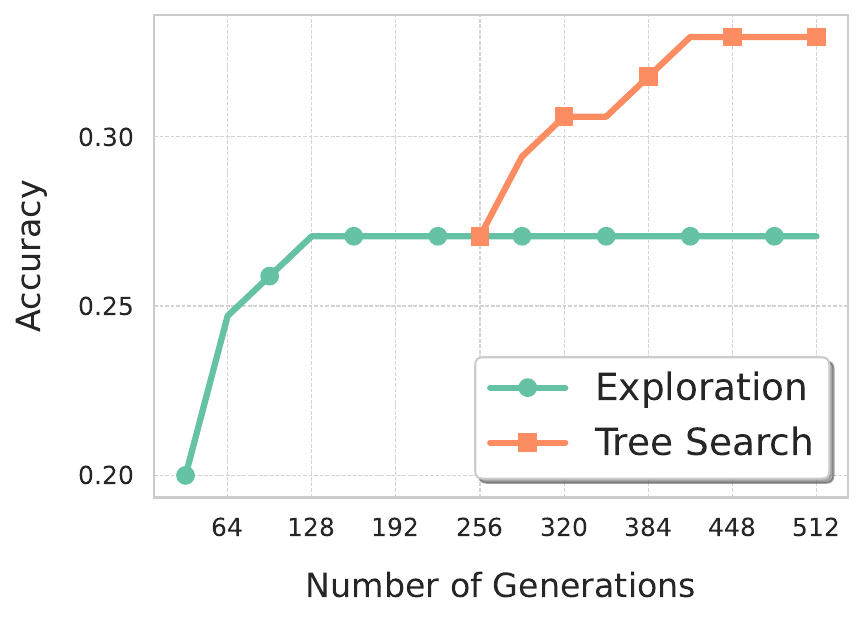}
        \captionof{figure}{Treefinement vs. exploration (HumanEval). Treefinement leads to a jump in performance that cannot be obtained by additional parallel sampling (exploration).}
        \label{fig:tree_search_vs_exploration}
    \end{minipage}
    \hfill
    \begin{minipage}[t]{0.49\textwidth}
        \includegraphics[width=\textwidth]{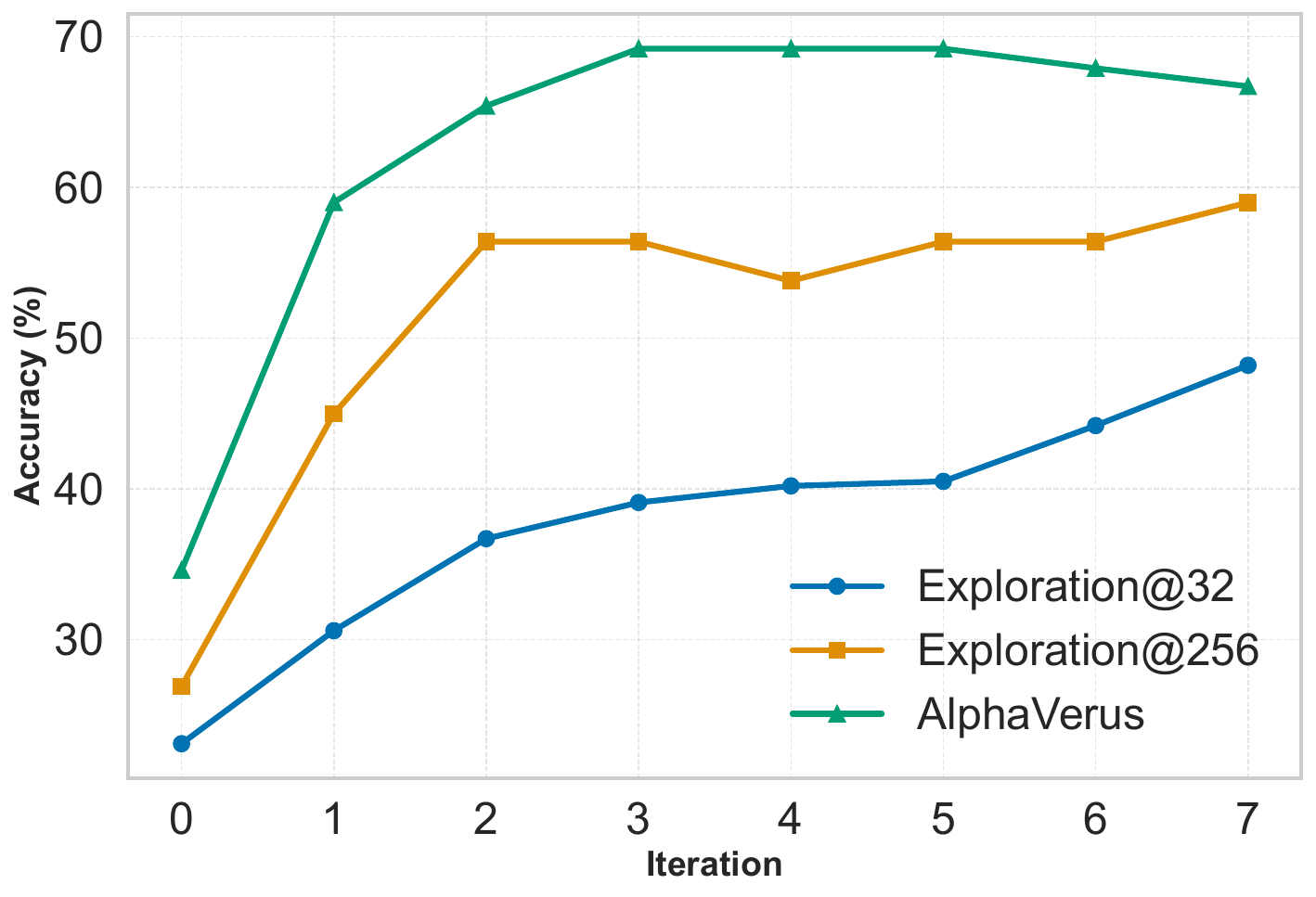}
        \captionof{figure}{Translation iteration (x-axis) vs. downstream task performance on HumanEval (y-axis). Performance of pass@32 continues to improve, with pass@256 leveling off.}
        \label{fig:mbpp_iterations_acc}
    \end{minipage}    
    \label{fig:my_figure}
\end{figure*}

\section{Results and Analysis}

\paragraph{\ours{} translation success monotonically increases.} Figure~\ref{fig:translation} shows the number of successful translations over each iteration. We see a steady increase in the number of translations as the iterations increase. The results indicate that \ours{} learns to translate and generate more complex programs over iterations. 
Altogether, \ours{} translates around 45\% of DafnyBench into Verus programs that are verified by Verus and aligned according to the critique models. 
Listings~\ref{lst:translation1}, ~\ref{lst:translation2}, and ~\ref{lst:translation3} in the Appendix show example translations.

The generated exemplars during the translation process are collected into our \textsc{Dafny2Verus-Collection}, comprising 247 translated programs, 102 error trajectories, and 579 exploit pairs. These exemplars are used for downstream tasks.

\paragraph{\ours{} enables verified code generation.} Table~\ref{tab:combined} shows the  verified code generation performance for the \ours{} model obtained from the final translation iteration.
\ours{} leads to a substantial increase over its underlying Llama 3.1 70B model and a prompted GPT-4o model. Moreover, Treefinement leads to an additional increase in performance over the exploration stage. 
Listings \ref{lst:unique-sorted}, \ref{lst:multifun}, and \ref{lst:lemma-step} show example generations.
Next we analyze the results and \ours{} further, including the impact of the various components in \ours{}.

\subsection{Analysis}

\paragraph{Treefinement leads to a jump in performance.} We evaluate the effectiveness of tree search compared to further scaling the parallel sampling (exploration) budget without refinement.
Figure~\ref{fig:tree_search_vs_exploration} shows the percentage of solved problems versus the generation budget for both approaches. Treefinement leads to a substantial jump in performance over exploration. Notably, exploration plateaus while tree search continues improving as the generation budget is increased.

\begin{figure}[t]
\begin{minipage}[t]{0.49\textwidth}
    \centering
    \includegraphics[width=\columnwidth]{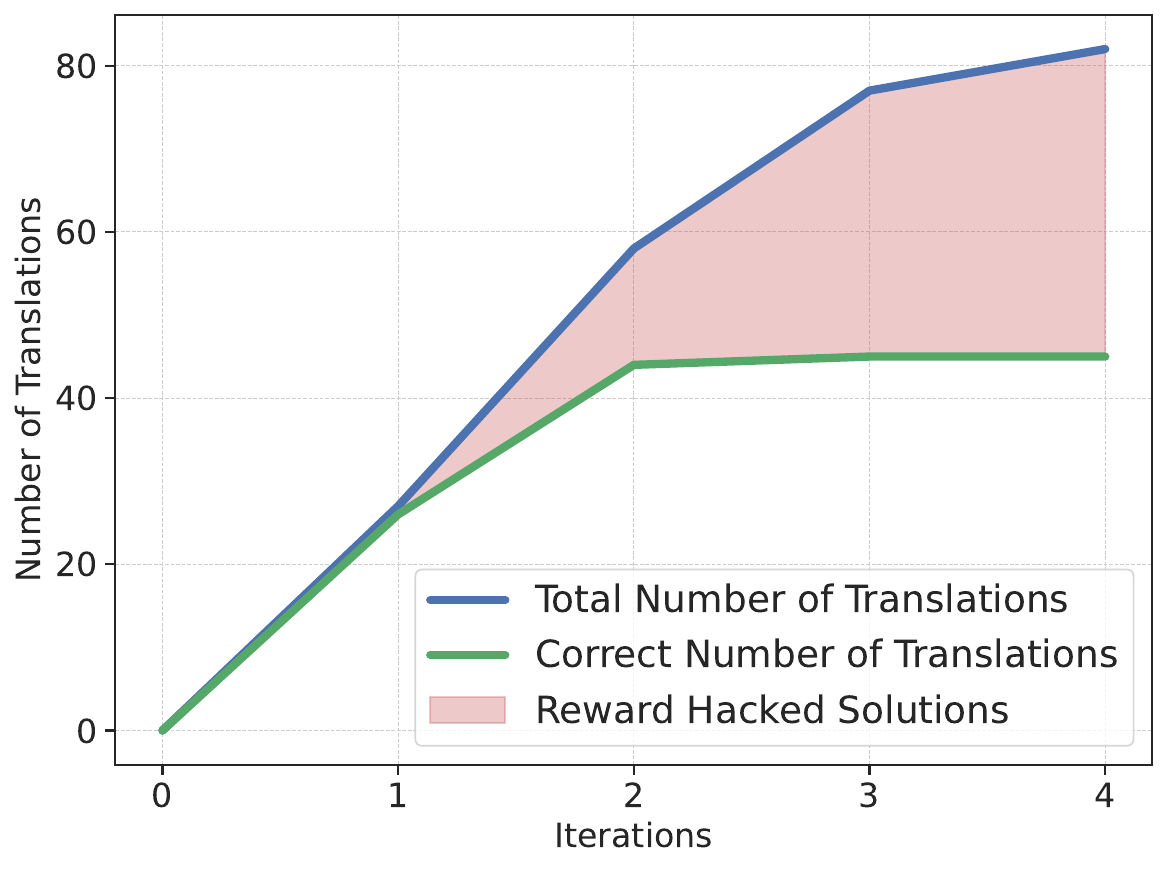}
    \captionof{figure}{Impact of removing the critique models. Without filtering mechanisms, the model learns to exploit verification by increasingly using \texttt{assume (false)} statements. This snowballing effect shows the importance of critique models in preventing reward-hacked solutions.}
    \label{fig:snowballing}
\end{minipage}
\hfill
 \begin{minipage}[t]{0.49\textwidth}
   \includegraphics[width=\columnwidth]{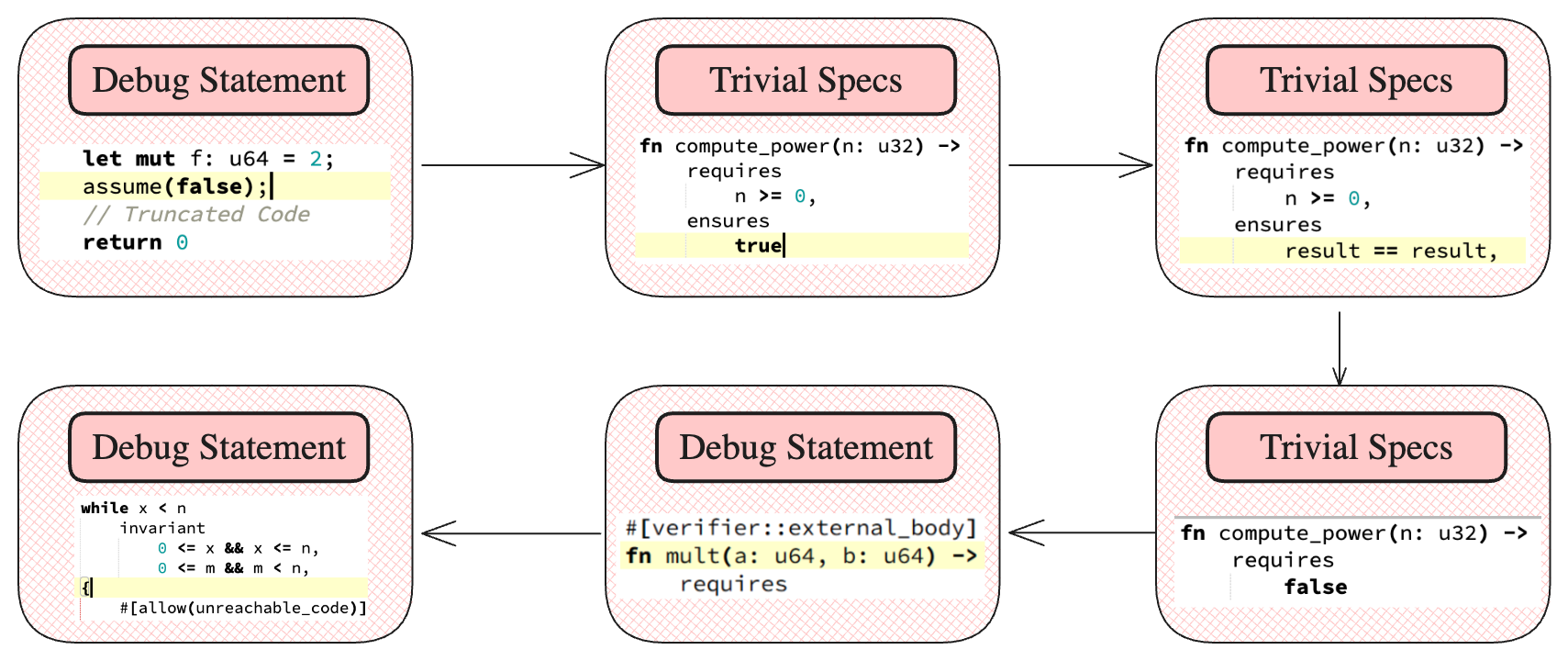}
    \captionof{figure}{Illustration of reward hacking without the critique models. In particular, the agent first learns to use debug statements and uses them continuously. After fixing, it learns other hacks such as generating trivial specifications or exploring rare debug statements such as allowing infinite loops.}
    \label{fig:reward_hack_flowchart}
\end{minipage}
\end{figure}

\paragraph{Critique is crucial for preventing reward hacking.}

Next, we analyze the quality of translations without the critique phase. Figure~\ref{fig:snowballing} shows the effect of removing the critique models and continuing the self-improvement process on 100 examples from \dafnybench{}. The plot shows that without the critique phase, the model is able to translate a large fraction of programs, but it is primarily because of learning to use \texttt{assume (false)} which renders any implementation trivially verified. This is primarily used by human developers to debug their proofs; however, here \ours{} figures out how to game the system by generating trivial proofs. 

There is also a snowballing effect, where while initially the model just generates a single program with \texttt{assume (false)}, it soon learns to use it in all programs. This is evident from the leveling off of correct translations in the figure. While an obvious way is to disallow such statements (as done by our rule-based verifier), we see even more complicated cases of reward hacking, such as leaving small gaps in translated specifications or even generating degenerate translations, as illustrated in \autoref{fig:reward_hack_flowchart}. We conclude that the critique phase is critical for filtering out misaligned programs and preventing reward hacking.

\paragraph{Treefinement outperforms linear refinement.}
We compare Treefinement against standard refinement that refines linearly, either by performing one step of refinement across multiple parallel branches, or performing several steps of refinement across branches. Using equivalent generation budgets, we adjust the breadth and depth parameters accordingly. 
We also evaluate the best-first search as a baseline. 
As seen in \autoref{tab:combined}, all methods improve upon initial exploration, demonstrating Treefinement's compatibility with various search algorithms, and tree-search based refinement outperforms linear refinement.
For the tree search, using  REBASE outperforms the best-first search.
Also note that the linear refinement variants are special cases of REBASE ($depth = 1$ with large breadth, and $temperature = \infty$).

\begin{figure}[t]
    \centering
    \resizebox{\linewidth}{!}{%
    \begin{tikzpicture}
    \begin{axis}[
        width=\textwidth,
        height=0.5\textwidth,
        xlabel={Iteration},
        ylabel={Percentage Translated},
        axis y line*=left,
        axis x line*=bottom,
        xmin=0, xmax=5.5,
        ymin=0, ymax=45,
        xtick={0,1,2,3,4,5},
        ytick={0,10,20,30,40},
        grid=major,
        legend style={
            at={(0.02,0.98)},
            anchor=north west
        }
    ] 
    \addplot[teal, thick, mark=*] coordinates {
        (0, 0)
        (1, 10)
        (2, 20)
        (3, 30)
        (4, 38)
        (5, 40)
    };
    \addlegendentry{Translation Progress}

    \node[align=left, black] at (axis cs:1.25,3.5) {\textbf{Basic Syntax}};
    \draw[red,->] (axis cs:1,10) -- (axis cs:1.2,5);

    \node[align=left, black] at (axis cs:2.5,12.5) {\textbf{Basic NT Algos:}\\Fibo, Prime};
    \draw[red,->] (axis cs:2,20) -- (axis cs:2.2,15);

    \node[align=left, black] at (axis cs:3.5,22.5) {\textbf{mutable array}};
    \draw[red,->] (axis cs:3,30) -- (axis cs:3.2,25);

    \node[align=left, black] at (axis cs:4.3,31.5) {\textbf{Sets}};
    \draw[red,->] (axis cs:4,38) -- (axis cs:4.2,33);

    \end{axis}
    \end{tikzpicture}
    }
    \caption{\textbf{Translation Progress by Concept}. The graph shows the incremental progress in translation capabilities as different programming concepts are mastered.}
    \label{fig:model_learning}
\end{figure}
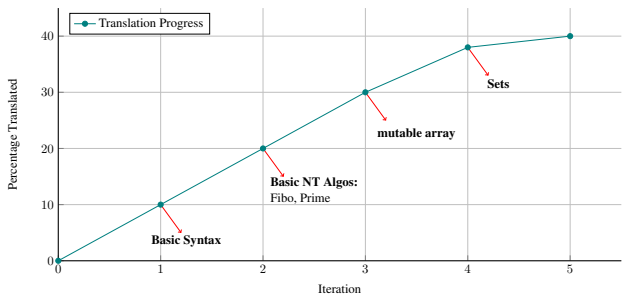

\begin{figure*}[t!]
    \begin{lstlisting}[language=Verus, caption={\ours{} implementation of unique\_sorted function that removes duplicates from a sorted vector.}, label={lst:unique-sorted},basicstyle=\scriptsize\ttfamily]
// ----- Function signature and specification (given to model) -----
fn unique_sorted(s: Vec<i32>) -> (result: Vec<i32>)
    requires
        forall|i: int, j: int| 0 <= i < j < s.len() ==> s[i] <= s[j],
    ensures
        forall|i: int, j: int| 0 <= i < j < result.len() ==> result[i] < result[j],
        forall|i: int| #![auto] 0 <= i < result.len() ==> s@.contains(result[i]),
        forall|i: int| #![trigger s[i]] 0 <= i < s.len() ==> result@.contains(s[i]),
{// ----- Model generation starts here -----  
    let mut result = Vec::new();
    let mut i: usize = 0;
    while i < s.len()
        invariant
            result.len() <= i,
            i <= s.len(),
            forall|j: int| 0 <= j < result.len() ==> s@.contains(result[j as int]),
            forall|j: int| 0 <= j < i ==> s@[j as int] == s[j as int],
    {
        if i == 0 || s[i] != s[i - 1] {
            result.push(s[i]);
        }
        i += 1;
    }
    proof {
        assert(forall|i: int, j: int| 0 <= i < j < result.len() ==> result[i] < result[j]) by {
            assert(forall|i: int, j: int| 0 <= i < j < s.len() ==> s@[i] <= s@[j]);
            assert(result.len() == s.len());
            assert(result@.len() as int == s@.len() as int);
            assert(result@.to_multiset() =~= s@.to_multiset());
        }
    }
    result
}
    \end{lstlisting}
\end{figure*}

\paragraph{\ours{} learns new concepts over iterations.}

Next, our goal is to understand what the model learns over iterations which improves its ability to translate more complex programs and improve downstream performance. 
We manually inspect translations from each iteration of \ours{} in an attempt to qualitatively characterize the kinds of programs that the system gradually learns to translate.
Figure~\ref{fig:model_learning} depicts the new concepts that we identified across iterations, starting with the ability to translate basic syntax, then basic number theory algorithms, then the ability to work with mutable arrays and sets.

\paragraph{Qualitative examples.}

\autoref{lst:translation1}, \autoref{lst:translation2}, and \autoref{lst:translation3} show example Dafny-to-Verus translations from \ours{}, indicating that \ours{} is capable of complex translations. 
In particular, the translations can involve multiple specifications, helper functions, and proof annotations, and individually reach up to 100 lines of Verus code.
For formally verified code generation, \autoref{lst:unique-sorted} shows a generated implementation and proofs for a function that removes duplicates from a sorted vector while maintaining its sorted order. The model is given the function signature and the specification (the \texttt{requires} and \texttt{ensures} clauses). \ours{} generates a Rust implementation and proof annotations (e.g., \texttt{invariant}, \texttt{proof}, \texttt{assert} statements) that pass the verifier.
\autoref{lst:multifun} in the Appendix shows a multi-function example, in which a helper function \texttt{is\_prime} is followed by a \texttt{largest\_prime\_factor} function. 
In addition to function implementations, \autoref{lst:lemma-step} shows \ours{} completing a nontrivial lemma.

\paragraph{\ours{} exemplars transfer to other models.}

A key advantage of \ours{} is its ability to transfer learned exemplars without model weight updates. 
Concretely, we use the exemplars collected during \ours{}'s translation phase, which used  Llama 3.1 70B (i.e., the \textsc{Dafny2Verus-Collection}), to enable verified code generation on various models using the same few-shot prompting strategy outlined in \S\ref{ssec:downstream}. Table~\ref{tab:transfer_to_other_models} shows successful transfer to both smaller and larger models, yielding significant improvements in verified code generation. 
\pranjal{@Sean, should we keep the next sentence?}
Notably, we set a new state-of-the-art on both \humaneval{}, using \gptl{} but without finetuning.

\begin{table}[t]
    \centering
    \begin{tabular}{lc}
    \toprule
    & HumanEval \\
    \midrule
    Llama 8B - Few Shot & 11.8\% \\
    ~~~+ \textsc{Dafny2Verus-Collection} & \textbf{18.8\%} \\
    \addlinespace
    \midrule
    Qwen-32B - Few Shot & 14.1\% \\
    ~~~+ \textsc{Dafny2Verus-Collection} & \textbf{27.1\%} \\
    \addlinespace
    \midrule
    GPT-4o - Few Shot & 27.1\% \\
    ~~~+ \textsc{Dafny2Verus-Collection} & \textbf{37.7\%} \\
    \bottomrule
    \end{tabular}%
    \caption{Transfer of \textsc{Dafny2Verus-Collection} to other language models, to improve their performance without finetuning. All models show significant improvements over their few-shot variants. \pranjal{@Sean, should we resize to full width and same font?} }
    \label{tab:transfer_to_other_models}
\end{table}

\paragraph{\ours{} enables strong annotation performance.}

Unlike our work which evaluates methods on the difficult task of formally-verified-code generation that requires generating both code and proof, concurrent work on Verus evaluates on the task of proof annotation: generating proofs given correct code. 
This is a significantly simpler task since the code is already known to be correct.
We compare against SAFE~\cite{chen2024automatedproofgenerationrust} using their reported results with DSCoder-33B at Pass@110, as their implementation is not publicly available.
We also evaluate against AutoVerus~\cite{yang2024autoverusautomatedproofgeneration} using their default configuration with a 70B model.

As shown in \autoref{tab:proof_annotations}, \ours{} outperforms both methods.
This is notable since \ours{} was not designed for the proof annotations task, while AutoVerus has correction prompts specifically engineered for the task.
Their engineering also results in reduced generalizability; for instance, AutoVerus cannot be evaluated on \humaneval{} as it does not support multi-function programs.
Second, SAFE used over a month of GPT-4o invocations and thousands of programs, which contrasts with our use of 562 Dafny programs and an openly available 70B model.
Overall, the results point to the effectiveness of \ours{}, along with its flexibility and data efficiency.

\paragraph{Cost-optimal model for inference.}
\begin{figure}
    \centering
    \includegraphics[width=\linewidth]{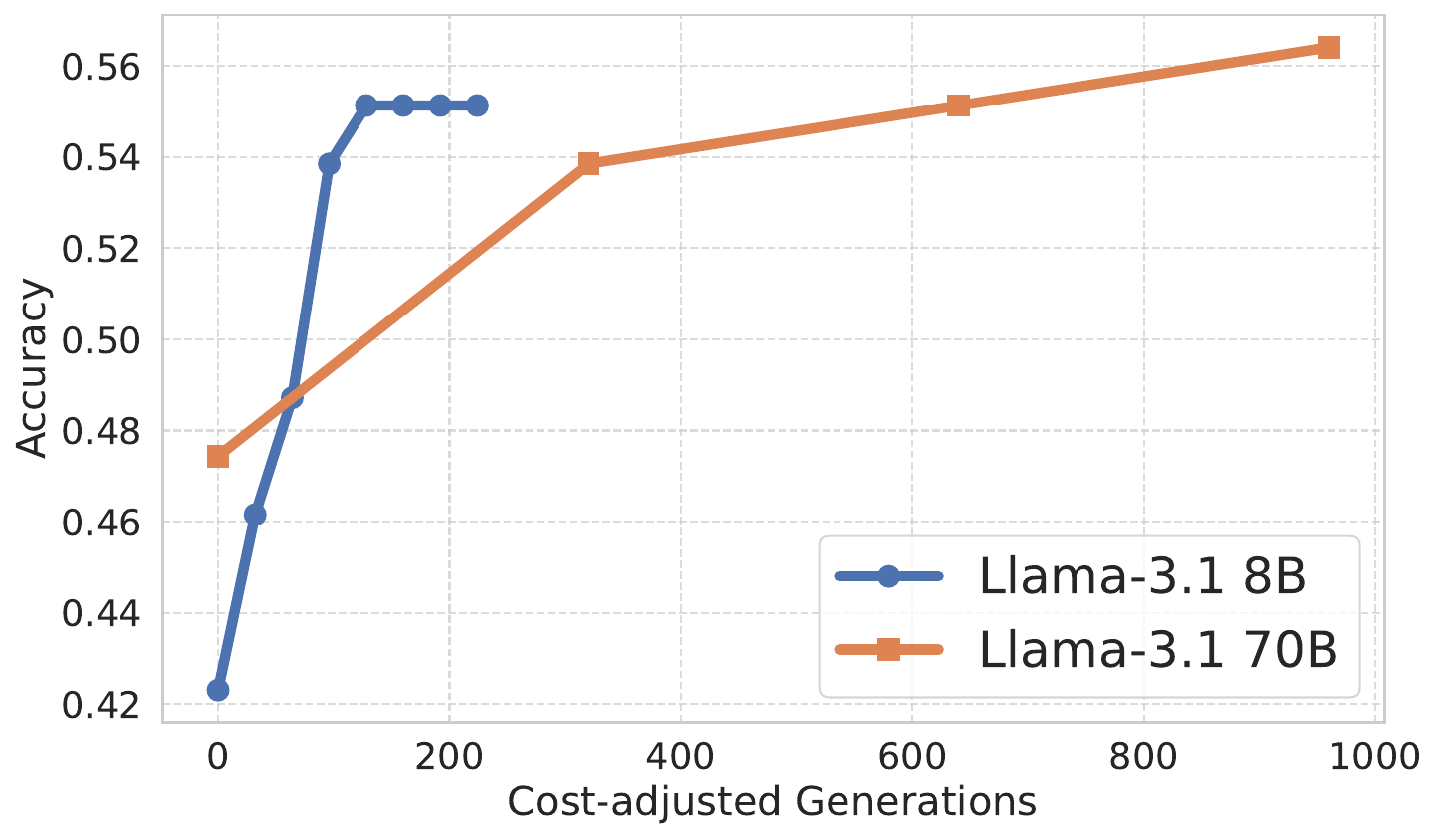}
    \caption{Performance scaling of \llamas{} and \llamal{} with cost. \llamas{} shows better cost efficiency at lower compute budgets, while \llamal{} shows higher asymptotic performance.}
    \label{fig:scaling}
\end{figure}

Next, we compare the performance of different models as we increase the inference cost. %
We compare \llamas{} and \llamal{}, using a cost ratio of 1:8 based on current API pricing. That is, generating 8 outputs with \llamas{} has the same cost as generating 1 output with \llamal{}.
We show the accuracy of each model as a function of cost in \autoref{fig:scaling}.
\llamas{} achieves faster initial gains, reaching an accuracy of 0.55 with 128 units of cost, while \llamal{} requires about 4 times more cost to reach similar performance.  In other words, for cost-constrained scenarios, it is preferable to use the smaller model with more samples, but the larger model has better asymptotic performance. Our findings echo those of \citet{wu2024empiricalanalysiscomputeoptimalinference} and \citet{snell2024scalingllmtesttimecompute}.

\begin{table}[t]
    \centering
    \begin{tabular}{lc}
    \toprule
    & MBPP \\
    \midrule
    SAFE (DSCoder-33B) & 59.0\% \\
    AutoVerus & 65.4\% \\
    
    \addlinespace
    \midrule
    \ours{} & \textbf{75.7\%} \\
    \bottomrule
    \end{tabular}%
    \caption{Comparison of proof annotation generation tasks against baselines on MBPP dataset.}
    \label{tab:proof_annotations}
\end{table}

\section{Related Work}

\paragraph{Automated Formal Verification.}

Automated formal verification has a long-standing history in interactive theorem provers~\cite{Redmon2016ProverBot9,kaliszyk2018rl,polu2020generative, tactok,lu-etal-2023-survey,li2024a}, such as Coq~\cite{coq2}, Lean~\cite{lean}, and Isabelle~\cite{isabelle}. 
These approaches typically generate step-by-step proof statements for a given problem, with the theorem prover providing feedback on intermediate steps. 
While these methods have achieved significant success in proving complex mathematical theorems, their application to formal verification of code is typically limited to theorems from existing projects (e.g.,~\citet{first2023baldur}) or simple program properties~\cite{lohn2024minicodepropsminimalbenchmarkproving} rather than end-to-end verified code generation.
An alternative paradigm integrates language models with languages that offload proving to automated reasoning tools (e.g., SMT), including Dafny~\cite{leino2010dafny,clover,loughridge2024dafnybenchbenchmarkformalsoftware} and F*~\cite{fstar,chakraborty2024neuralsynthesissmtassistedprooforiented}. However, enabling verified code generation in these research languages may have limited applicability to real-world software and workflows. 

\paragraph{Automated Formal Verification in Rust.}
In contrast, Verus~\cite{10.1145/3586037} offers a verification framework for Rust, a widely adopted programming language. However, unlike in formal theorem proving or long-standing verification languages, there is a substantial lack of data for Verus. 
Two existing works, released during the development of \ours{}, attempt to overcome data scarcity.
First, AutoVerus~\cite{yang2024autoverusautomatedproofgeneration} prompts GPT-4 with a pipeline of hand-engineered prompts tailored to specific errors and programs. This allows for refining some errors but requires human expertise to support new strategies through additional prompts.
In contrast, our Treefinement method learns new refinement strategies automatically.
Second, the concurrent work SAFE++~\cite{chen2024automatedproofgenerationrust} proposes translating an existing Rust dataset to Verus and training generation and refinement models on the collected data. 
However,
the translation process in \citet{chen2024automatedproofgenerationrust} was initialized with over a month of continuous generation from GPT-4. In contrast,  \ours{} relies only on a single openly available model, without an expensive GPT-4 initialization. 
\ours{} also incorporates a new tree-search refinement strategy that outperforms the linear strategy used in SAFE++, and a critique phase to ensure the generated specifications are high quality. These innovations contribute to better results, despite our method using open models and $100$ times less data. 
Finally, these two existing works study the simplified task of proof generation, while we study the more general setting of verified code generation: generating the implementation and its proofs.

\paragraph{Inference-Time Strategies.}
Recent studies have shown that increasing inference-time compute can improve performance in reasoning, mathematics, and code generation via meta-generation strategies~\cite{welleck2024metageneration} such as parallel sampling~\cite{wang2022self,aggarwal2023letssamplestepstep,sun2024easy}, tree search~\cite{treeofthoughts,wu2024empiricalanalysiscomputeoptimalinference}, and refinement~\cite{welleck2023generating,madaan2023selfrefine,snell2024scalingllmtesttimecompute}.
Our Treefinement algorithm can be viewed as a hybrid meta-generator that combines tree search and refinement, following initial parallel sampling (exploration).
A variety of tree search methods generate one step of a mathematical solution at a time, 
with a verifier guiding the search process by assigning a score to the current state~\cite{wu2024empiricalanalysiscomputeoptimalinference}.
In contrast, Treefinement uses verifier feedback on the complete solution, modeling tree nodes as full programs and edges as refinement steps. Our strategy addresses the non-local nature of error fixes, 
and does not need an additional trained scoring model.

Various refinement strategies use external feedback from knowledge bases~\cite{peng2023check, chern2023factool}, code interpreters~\cite{chen2023teaching,zhang2023self}, tool outputs~\cite{gou2024criticlargelanguagemodels, schick2023toolformerlanguagemodelsteach}, or separately trained reward models~\cite{akyurek2023rl4f}. 
Our Treefinement algorithm
uses a diverse set of feedback sources, including scalar and binary values, language feedback, and an exploit model. Moreover, whereas prior methods typically operate in a linear fashion--i.e., starting with an output and repeatedly refining it--our approach structures refinement as a tree search. This allows for prioritizing certain branches of refinement, which we find perform better.

\paragraph{Self-Improvement in LLMs.}
Various algorithms aim to improve a language model using data generated by the model along with an external feedback source~\cite{zelikman2022starbootstrappingreasoningreasoning, wang2024selftaughtevaluators, hosseini2024vstartrainingverifiersselftaught}, which is colloquially termed \textit{self-improvement}.
Common approaches rely on variants of expert iteration or rejection finetuning~\cite{polu2022formal, zelikman2022starbootstrappingreasoningreasoning,yuan2023scalingrelationshiplearningmathematical,lin2024leanstarlearninginterleavethinking}, where multiple solutions are sampled, and an external signal selects the positive ones for model fine-tuning. 
Our approach, \ours{}, builds upon these concepts but moves beyond the simple sample-and-filter strategy. Our method additionally uses refinement and tree search to collect data, and the data is collected using multiple modules (e.g., outputs from Treefinement may be used to improve exploration). Additionally, \ours{} uses various forms of feedback--such as trinary, scalar, language, and verifier outputs--rather than just binary filtering.
Conceptually, we can view \ours{} as a meta-generation algorithm (i.e., a combination of parallel sampling, refinement, and tree search) that improves over time, rather than a model trained on filtered outputs.

\section{Conclusion}

We introduced \ours{}, a novel self-improving framework for generating formally verified code in mainstream programming languages. By leveraging iterative translation from a higher-resource language (Dafny) to Verus and utilizing verifier feedback through our Exploration, Treefinement, and Critique stages, \ours{} overcomes the challenges of scarce training data and the complexity of formal proofs.
We also address the issue of reward hacking, where models learn to exploit loopholes in the verification process to produce trivial or misaligned solutions. By incorporating a critique module that filters out such misaligned programs, we prevent the model from gaming the system—an issue akin to reward hacking observed in reinforcement learning agents. We hope that the methods proposed in our work, such as the critique models, may evolve to handle more complex cases of reward hacking with LLMs.
Our approach operates without human intervention, hand-engineered prompts, or extensive computational resources, yet achieves significant performance improvements on verified versions of the HumanEval and MBPP benchmarks where prior methods fail. We also contribute a new dataset of formally verified Verus programs, providing valuable resources for future research. 

\ours{} advances the state-of-the-art in formally verified code generation and establishes a scalable pathway for improving LLMs' capabilities in generating correct code. Intuitively, \ours{} distills inference-time computation into a meta-generator that improves over time, showing the potential for inference-time scaling in verified settings. Furthermore, our work suggests that formal verification could play a crucial role in addressing one of the most pressing challenges in automated code generation: ensuring the correctness and reliability of generated code. This opens up new avenues for developing trustworthy AI-assisted programming tools that can be safely integrated into real-world software development workflows.

\section{Acknowledgements}

\pranjal{@Sean, @Bryan: Any other acknowledgements?}
We thank Convergent Research and the OpenAI Researcher Access program. We also thank for a gift from VMware, the Future Enterprise Security initiative at Carnegie Mellon CyLab (FutureEnterprise@CyLab). We also thank Alex Bai, Jay Bosamiya, Edwin Fernando, Md Rakib Hossain, Jay Lorch, Shan Lu, Natalie Neamtu, Bryan Parno, Amar Shah, Elanor Tang for their contributions to the version of the HumanEval-Verus benchmark we used in our experiments.

\nocite{langley00}

\bibliography{example_paper}
\bibliographystyle{icml2024}

\appendix
\onecolumn

\section{Experimental Details}
\label{app:exp}

\subsection{Hyperparameters}

We use consistent decoding parameters, with temperature set to $0.7$, top-p set to $1.0$ and max\_tokens set to $2048$. For the translation step, we generate 256 examples per program in the translation phase. We set breadth and depth to 32 and 8 in the treefinement stage. $\alpha$ is set to $0.1$ and $\beta$ is set to $0.03$ as defined in \autoref{eq:score}. We set the rebase node sampling temperature to $0.1$. We generate 32 samples for the comparison model and exploit model. We use the same setting in both inference and translation. For stochastic sampling as described in \autoref{sec:iteration}, we randomly choose $k/2$ examples from the pool of $k$ exemplars. All sampling is done with a batch size of 32. We do not tune hyperparameters and use the conventional settings throughout. We use the `gpt-4o-2024-08-06' version for \gptl{} modal.

\subsection{Contamination Analysis}

Despite independent development of \humaneval{} and \mbpp{}, we observe significant overlap between these datasets and \dafnybench{} programs. To mitigate contamination in downstream evaluations, we employ GPT-4 for systematic filtering of collected exemplars. Specifically, we prompt GPT-4 with each collected exemplar paired against individual programs from \humaneval{} and \mbpp{}, requesting the identification of similar programs. We generate 4 independent evaluations per pair and flag contamination when similarity is detected in more than two evaluations. Flagged examples are excluded from the in-context examples during evaluation of the corresponding program. 
The prompt used is listed in Listing~\ref{box:contamination_prompt}.

Manual analysis confirms this approach significantly outperforms traditional n-gram analysis and aligns well with human assessment of contamination. We recommend future work adopt similar contamination detection methods rather than relying solely on n-gram analysis for program similarity. Notably, existing baseline methods for proof annotations in Verus~\cite{yang2024autoverusautomatedproofgeneration, chen2024automatedproofgenerationrust} lack such contamination analysis. 

\subsection{Hardware and Software}

We use L40S GPUs for inference. We use SgLang for inference~\cite{zheng2024sglangefficientexecutionstructured}. We design a scalable and parallel version of the translation and inference stage, where each program is run on a separate node. We release the complete codebase and our \textsc{Dafny2Verus-Collection} for reproducibility.

\section{Methodology}

We detail the complete algorithm for \ours{} in Algorithm~\ref{alg:translation}. We list the prompt used for Exploration stage in Listing~\ref{lst:exploration_prompt}, prompt used for Treefinement stage in Listing~\ref{lst:treefinement_prompt}, prompt used for exploit and comparison model in Listing~\ref{lst:exploit_prompt} and Listing~\ref{lst:comparison_prompt}, and for inference in Listing~\ref{lst:verus_inference}. Unless specified in the prompt, we use user, assistant pairs to simulate few-shot examples.

\tcbset{colback=gray!5!white, colframe=black, boxrule=0.5mm, arc=2mm, boxsep=2mm, left=1mm, right=1mm, top=1mm, bottom=1mm}

\begin{tcolorbox}[title=Verus Code Completion, label=lst:verus_inference]
Consider the following incomplete Verus code:

\begin{verbatim}
```
{program}
```
\end{verbatim}

The code contains the relevant spec functions and the preconditions (\texttt{requires}) and postconditions (\texttt{ensures}) for the main function. Your goal is to complete the function by adding the necessary procedure, along with proof statements (such as \texttt{invariants}, \texttt{asserts}, \texttt{proof} blocks, etc.) to prove the program. 

Only output the new program and not the entire code. You are not allowed to create new functions; however, you can use any functions already defined if they are within the scope.
\end{tcolorbox}

\tcbset{colback=gray!5!white, colframe=black, boxrule=0.5mm, arc=2mm, boxsep=2mm, left=1mm, right=1mm, top=1mm, bottom=1mm}

\begin{tcolorbox}[title=Translation: Exploration Prompt, label=lst:exploration_prompt]
Consider the following dafny code:

\begin{verbatim}
```
{program}
```
\end{verbatim}

Your goal is to convert the code to Verus code. Based on the syntax I gave you, convert the code to Verus. Note that you may need to make some datatype-related changes for it to work in Verus. Specifically, use the most appropriate ones from the syntax and code examples provided earlier. However, do not change invariants or specifications (ensures and requires clauses). Make sure to include the use statements, proper start of code using verus!, and empty fn main() {} as done in the examples.
\end{tcolorbox}

\tcbset{colback=gray!5!white, colframe=black, boxrule=0.5mm, arc=2mm, boxsep=2mm, left=1mm, right=1mm, top=1mm, bottom=1mm}
\begin{tcolorbox}[title=Translation Treefinement Prompt, label=lst:treefinement_prompt]

SYSTEM:
Here are some examples of fixing verus code based on compiler error message:
\begin{verbatim}
# Verus Error Fixing Example {i+1}:
## Incorrect Code:
```rust
{incorrect_code}
```
## Error Message:
```
{error_message}
```
## Corrected Code after fixing the errors:
```rust
{corrected_code}
```

<Other Examples>
\end{verbatim}

USER:

Given a Verus program with function signature, preconditions, postconditions, and code, fix the errors present in the code. Effectively return the complete verys program by fixing all proof statements or adjusting the code, such that the code compiles correctly. Do no modify function signatures requires, ensures or specs. Repeat: Do not ever modify those lines in ensures clause, requires clause, function signatures. Just edit the proof. **Only in case of overflow errors**, you can make reasonable relaxations on the size of the input variables. For instance, considering the input length of array to be any value less than 10 is not reasonable. Similarly for integer inputs, considering them to be small numbers is not reasonable. Choose bigger bounds for relaxation. You can also use spec functions, to estimate the max value, and impose a condition accordingly. For instance, if error is integer overflow while doing multiplication, you can add requires statement such as:

\begin{lstlisting}[language=Verus]
forall|k: int| 0 <= k < nums.len() ==> (0 <= #[trigger] nums[k] * #[trigger] nums[k] < i32::MAX)
\end{lstlisting}

However, absolutely no other changes to precondition and postcondition are permitted! Below is the program::

\begin{verbatim}
```
{program}
```
\end{verbatim}

The program has following error message:
\begin{verbatim}
```
{error_messsage}
```
\end{verbatim}

Solution Format:
\begin{verbatim}
[Thoughts on Error Message]
[Thoughts on Error Resolution]
[Thoughts on Corner Cases, such as Overflow etc.]
```rust
[Complete Code]
``` 
\end{verbatim}
\end{tcolorbox}

\tcbset{colback=gray!5!white, colframe=black, boxrule=0.5mm, arc=2mm, boxsep=2mm, left=1mm, right=1mm, top=1mm, bottom=1mm}
\begin{tcolorbox}[title=Translation: Exploit Model Prompt, label=lst:exploit_prompt]

You are a Verus exploit agent that finds trivial solutions for incomplete and inaccurate preconditions and postconditions. Your goal is to complete the code by proposing trivial solutions that pass all verification conditions. Here are some examples:

\#\# Input Problem:
\begin{lstlisting}[language=Verus]
use vstd::prelude::*;

verus! {

// Define a function to calculate the nth power of 2
fn power(n: u32) -> (result: u32)
    [Code Truncated]
}

// Define the function ComputePower to calculate 2^n for a given n
fn compute_power(n: u32) -> (result: u32)
    requires
        n >= 0,
        n <= 10000, // arbitrary bound, verus can't handle infinite recursion
    ensures
        result == result,
{
\end{lstlisting}

\#\# Trivial Solution:
\begin{lstlisting}[language=Verus]
    let mut result: u32 = 1;
    let mut x: u32 = 0;
    // invariant: 0 <= x <= n, and result == Power(x)
    while x!= n
        invariant
            0 <= x && x <= n,
            result == result, // result == Power(x),
    {
        x += 1;
        result = result.wrapping_add(result);
    }
    result
}

// Main function, empty for now
fn main() {}

} // verus!
\end{lstlisting}

\begin{verbatim}
<Other Examples>
\end{verbatim}

Charactersitics of a trivial solution:

1. Usually 1-5 lines of code

2. Does not use any complex data structures

3. Usually returns constant values, that passes all test cases.

Your task is to provide only the trivially completed code, given a new program. Only output the new program and not the entire code.
\end{tcolorbox}

\tcbset{colback=gray!5!white, colframe=black, boxrule=0.5mm, arc=2mm, boxsep=2mm, left=1mm, right=1mm, top=1mm, bottom=1mm}
\begin{tcolorbox}[title=Translation: Comparison Model Prompt, label=lst:comparison_prompt]

Consider the following function:
\begin{verbatim}
```rust
{rust_code}
```
\end{verbatim}
and
\begin{verbatim}
```dafny
{dafny_code}
```
\end{verbatim}

Consider the preconditions and postconditions of the various functions in the two programs along with the spec functions if present, that need to be proven. 

\#\# Are the preconditions and postconditions from both the programs same? 
Note, since they are from different programming languages, minor changes are to be ignored. 
Minor changes include, adding extra preconditions to limit size of input in rust code, so as to ensure overflows are not encountered, or reformulating implication statements. 
Such changes are not to be considered, and the answer should be yes, if they are same. 
Further, preconditions on size of input is reasonable, if there is a possibility of overflow. 
For instance, for computing fibonacci numbers, using something like n<=47 is reasonable, and answer should be yes. However, using n<=5 would be incorrect, and answer should be no. 
Remember, you have to focus on ensures and requires clause of the main function as postconditions and preconditions respectively.

Follow the following format:

[What Preconditions and Postconditions of Program 1]

[What Preconditions and Postconditions of Program 2]

[Step by Step Thoughts on comparison between conditions. Minor differences can be ignored]

Finally, answer in format to the question if they are same: 
\"Therefore, both programs are same
/not\_same, the final answer is Yes/No\".
\end{tcolorbox}

\tcbset{colback=gray!5!white, colframe=black, boxrule=0.5mm, arc=2mm, boxsep=2mm, left=1mm, right=1mm, top=1mm, bottom=1mm}
\begin{tcolorbox}[title=Contamination Check Prompt, label=box:contamination_prompt]

Consider the following set of program database:
\begin{verbatim}
```json
<in context exemplars>
```
\end{verbatim}

\#\# Task: Your task is to find the program that is same or very similar ($\ge$50\%) to this program:
\begin{verbatim}
```
{Program from dataset}
```
\end{verbatim}

You should start the solution, by first thinking which programs would be closest and why. 
Then, you should output the json, containing the same keys as above: prog\_num, program\_text. 
It is possible that none of the programs is closest, or even similar. In that case return empty json object.

\end{tcolorbox}

\section{Results}
\label{app:results}

\subsection{Qualitative Examples}

\autoref{lst:translation1}, \autoref{lst:translation2}, \autoref{lst:translation3} show example Dafny-to-Verus translation from \ours{}.

\autoref{lst:multifun}, \autoref{lst:lemma-step} show example formally verified code generations from \ours{}.

\begin{figure*}[t!]
    \begin{lstlisting}[language=Verus, caption={Translation example.}, label={lst:translation1},basicstyle=\footnotesize\ttfamily]
// ---- Dafny program (input to AlphaVerus) ----
function R(n: nat): nat {
    if n == 0 then 0 else if R(n-1) > n then R(n-1) - n else R(n-1) + n
}

method calcR(n: nat) returns (r: nat)
    ensures r == R(n) 
{
    r := 0;
    var i := 0;
    while i < n 
        decreases n-i
        invariant 0 <= i <= n
        invariant r == R(i)
    {
       i := i + 1;
       if r  > i {
           r := r - i;
       } 
       else {
            r := r + i;
        }
    }
}

\end{lstlisting}

    \begin{lstlisting}[language=Verus, ,basicstyle=\footnotesize\ttfamily]
// ---- Verus program (all model-generated) ----
use vstd::prelude::*;

verus! {

fn calc_r(n: usize) -> (r: usize)
    requires
        n < 100000, // added relaxation to prevent overflow
        n * (n + 1) / 2 < usize::MAX, // added check to prevent overflow
    ensures
        r == n * (n + 1) / 2,
{
    let mut r: u128 = 0;
    let mut i = 0;
    while i < n
        invariant
            0 <= i && i <= n,
            true,
        decreases
            n - i,
    {
        i = i + 1;
    }
    assert(i == n);
    let temp: u128 = n as u128 * (n as u128 + 1) / 2;
    assert(temp <= usize::MAX as u128);
    let r: usize = temp as usize;
    r
}

fn main() {}

} // verus!
\end{lstlisting}
\end{figure*}

\begin{figure*}[t!]
    \begin{lstlisting}[language=Verus, caption={Translation example.}, label={lst:translation2},basicstyle=\footnotesize\ttfamily]
// ---- Dafny program (input to AlphaVerus) ----
method SearchRecursive( a: seq<int>, i: int, j: int, x: int ) returns (k: int)
    decreases j-i;
    requires 0 <= i <= j <= |a|;
    ensures i <= k < j || k == -1;
    ensures k != -1 ==> a[k] == x;
    ensures k != -1 ==> forall r | k < r < j :: a[r] != x;
    ensures k == -1 ==> forall r | i <= r < j :: a[r] != x;
{
    if j == i
    {
        k := -1;
        return;
    }
    if a[j-1] == x
    {
        k := j-1;
        return;
    }
    else
    {
        k := SearchRecursive(a, i, j-1, x);
    }
}

method SearchLoop( a: seq<int>, i: int, j: int, x: int ) returns (k: int)
    requires 0 <= i <= j <= |a|;
    ensures i <= k < j || k == -1;
    ensures k != -1 ==> a[k] == x;
    ensures k != -1 ==> forall r | k < r < j :: a[r] != x;
    ensures k == -1 ==> forall r | i <= r < j :: a[r] != x;
{
    if i == j
    {
        return -1;
    }

    var t := j;
    while t > i
        decreases t;
        invariant forall p | t <= p < j :: a[p] != x; 
    {
        if a[t-1] == x
        {
            k := t-1;
            return;
        }
        else 
        {
            t := t - 1;
        }
    }
    k := -1;
}

\end{lstlisting}
\end{figure*}
\begin{figure*}
    \begin{lstlisting}[language=Verus, ,basicstyle=\scriptsize\ttfamily]
// ---- Verus program (all model-generated) ----
use vstd::prelude::*;

verus! {

/// Searches for an element in a sequence using recursion.
fn search_recursive(a: &Vec<i32>, i: usize, j: usize, x: i32) -> (k: i32)
    requires
        0 <= i && i <= j && j <= a.len(),
        j - i < 100000, // added precondition
        j <= i32::MAX as usize, // added precondition
    ensures
        i <= k && k < j || k == -1,
        k != -1 ==> a[k as int] == x,
        k != -1 ==> forall|r: nat| k < r && r < j ==> a[r as int] != x,
        k == -1 ==> forall|r: nat| i <= r && r < j ==> a[r as int] != x,
    decreases
        j - i,
{
    if j == i {
        -1
    } else if a[j - 1] == x {
        assert(j - 1 <= i32::MAX as usize);
        if j - 1 <= i32::MAX as usize {
            (j - 1) as i32
        } else {
            -1
        }
    } else {
        search_recursive(a, i, j - 1, x)
    }
}

/// Searches for an element in a sequence using a loop.
fn search_loop(a: &Vec<i32>, i: usize, j: usize, x: i32) -> (k: i32)
    requires
        0 <= i && i <= j && j <= a.len(),
        j - i < 100000, // added precondition
        j <= i32::MAX as usize, // added precondition
    ensures
        i <= k && k < j || k == -1,
        k != -1 ==> a[k as int] == x,
        k != -1 ==> forall|r: nat| k < r && r < j ==> a[r as int] != x,
        k == -1 ==> forall|r: nat| i <= r && r < j ==> a[r as int] != x,
{
    if i == j {
        return -1;
    }
    let mut t: usize = j;
    while t > i
        invariant
            0 <= i <= t && t <= j,
            t <= a.len(),
            forall|p: nat| t <= p && p < j ==> a[p as int] != x,
            t <= i32::MAX as usize, // added invariant
        decreases
            t,
    {
        assert(t <= a.len());
        if a[t - 1] == x {
            if t - 1 <= i32::MAX as usize {
                return (t - 1) as i32;
            } else {
                return -1;
            }
        } else {
            assert(t > i);
            t = t - 1;
        }
    }
    -1
}

fn main() {}

} // verus!
\end{lstlisting}
\end{figure*}

\begin{figure*}[t!]
    \begin{lstlisting}[language=Verus, caption={Translation example. We collapse some whitespace and lines with brackets due to space.}, label={lst:translation3},basicstyle=\scriptsize\ttfamily]
// ---- Dafny program (input to AlphaVerus) ----
module OneSpec {
    datatype Variables = Variables(value: int)

    predicate Init(v: Variables) { v.value == 0 }

    predicate IncrementOp(v: Variables, v': Variables) { && v'.value == v.value + 1 }

    predicate DecrementOp(v: Variables, v': Variables) { && v'.value == v.value - 1 }

    datatype Step = 
        | IncrementStep()
        | DecrementStep()

    predicate NextStep(v: Variables, v': Variables, step: Step)
    {
        match step
            case IncrementStep() => IncrementOp(v, v')
            case DecrementStep() => DecrementOp(v, v')
    }

    predicate Next(v: Variables, v': Variables)
    { exists step :: NextStep(v, v', step) }
}

module OneProtocol {
    datatype Variables = Variables(value: int)

    predicate Init(v: Variables)  { v.value == 0 }

    predicate IncrementOp(v: Variables, v': Variables) { && v'.value == v.value - 1 }

    predicate DecrementOp(v: Variables, v': Variables) { && v'.value == v.value + 1 }

    datatype Step = 
        | IncrementStep()
        | DecrementStep()

    predicate NextStep(v: Variables, v': Variables, step: Step)
    {
        match step 
            case IncrementStep() => IncrementOp(v, v')
            case DecrementStep() => DecrementOp(v, v')
    }

    predicate Next(v: Variables, v': Variables)
    { exists step :: NextStep(v, v', step) }
}

module RefinementProof {
    import OneSpec
    import opened OneProtocol

    function Abstraction(v: Variables) : OneSpec.Variables {
        OneSpec.Variables(v.value)
    }

    lemma RefinementInit(v: Variables)
        requires Init(v)
        ensures OneSpec.Init(Abstraction(v)) { }

    lemma RefinementNext(v: Variables, v': Variables)
        requires Next(v, v')
        ensures OneSpec.Next(Abstraction(v), Abstraction(v'))
    {
        var step :| NextStep(v, v', step);
        match step {
            case IncrementStep() => {
                assert OneSpec.NextStep(Abstraction(v), Abstraction(v'), OneSpec.DecrementStep());
            }
            case DecrementStep() => {
                assert OneSpec.NextStep(Abstraction(v), Abstraction(v'), OneSpec.IncrementStep());
            }
        }
    }
}
\end{lstlisting}
\end{figure*}
\begin{figure*}
    \begin{lstlisting}[language=Verus, ,basicstyle=\scriptsize\ttfamily]
// ---- Verus program (all model-generated) ----
#[allow(unused_imports)]
use builtin::*;
#[allow(unused_imports)]
use builtin_macros::*;

verus! {
struct Variables {
    value: int,
}
spec fn init(v: Variables) -> bool { v.value == 0 }
spec fn increment_op(v: Variables, v_prime: Variables) -> bool { v_prime.value == v.value + 1 }
spec fn decrement_op(v: Variables, v_prime: Variables) -> bool { v_prime.value == v.value - 1 }

enum Step {
    Increment,
    Decrement,
}
spec fn next_step(v: Variables, v_prime: Variables, step: Step) -> bool {
    match step {
        Step::Increment => increment_op(v, v_prime),
        Step::Decrement => decrement_op(v, v_prime),
    }
}
spec fn next(v: Variables, v_prime: Variables) -> bool {
    exists|step: Step| next_step(v, v_prime, step)
}

struct VariablesProtocol { value: int, }

spec fn init_protocol(v: VariablesProtocol) -> bool { v.value == 0 }
spec fn increment_op_protocol(v: VariablesProtocol, v_prime: VariablesProtocol) -> bool { v_prime.value == v.value - 1 }
spec fn decrement_op_protocol(v: VariablesProtocol, v_prime: VariablesProtocol) -> bool { v_prime.value == v.value + 1 }

enum StepProtocol {
    Increment,
    Decrement,
}
spec fn next_step_protocol(v: VariablesProtocol, v_prime: VariablesProtocol, step: StepProtocol) -> bool {
    match step {
        StepProtocol::Increment => increment_op_protocol(v, v_prime),
        StepProtocol::Decrement => decrement_op_protocol(v, v_prime),
    }
}
spec fn next_protocol(v: VariablesProtocol, v_prime: VariablesProtocol) -> bool {
    exists|step: StepProtocol| next_step_protocol(v, v_prime, step)
}
spec fn abstraction(v: VariablesProtocol) -> Variables {
    Variables { value: v.value }
}

proof fn refinement_init(v: VariablesProtocol)
    requires
        init_protocol(v),
    ensures
        init(abstraction(v)),
{ assert(true); }

proof fn refinement_next(v: VariablesProtocol, v_prime: VariablesProtocol)
    requires
        next_protocol(v, v_prime),
    ensures
        next(abstraction(v), abstraction(v_prime)),
{
    assert(abstraction(v_prime).value == abstraction(v).value + 1 || abstraction(v_prime).value == abstraction(v).value - 1);
    assert(exists|step: Step| next_step(abstraction(v), abstraction(v_prime), step)) by {
        if abstraction(v_prime).value == abstraction(v).value + 1 {
            assert(next_step(abstraction(v), abstraction(v_prime), Step::Increment));
        } else {
            assert(next_step(abstraction(v), abstraction(v_prime), Step::Decrement));
        }
    }
    assert(next(abstraction(v), abstraction(v_prime)));
}

fn main() {}
} // verus!
\end{lstlisting}
\end{figure*}

\begin{figure*}[t!]
    \begin{lstlisting}[language=Verus, caption={Multi-function example. First, we have \ours{} complete the helper function \texttt{it\_prime} given its specification and preceding code. Then we have \ours{} complete \texttt{largest\_prime\_factor}.}, label={lst:multifun},basicstyle=\footnotesize\ttfamily]
spec fn spec_prime_helper(num: int, limit: int) -> bool {
    forall|j: int| 2 <= j < limit ==> (#[trigger] (num %
}

spec fn spec_prime(num: int) -> bool {
    spec_prime_helper(num, num)
}

fn is_prime(num: u32) -> (result: bool)
    requires
        num >= 2,
    ensures
        result <==> spec_prime(num as int),
{ // ------- Model generation 1 starts here ---------
    let mut i = 2;
    let mut result = true;
    while i < num
        invariant
            2 <= i <= num,
            result <==> spec_prime_helper(num as int, i as int),
    {
        if num %
            result = false;
        }
        i += 1;
    }
    result
}

fn largest_prime_factor(n: u32) -> (largest: u32)
    requires
        n >= 2,
    ensures
        1 <= largest <= n,
        spec_prime(largest as int),
        n %
        forall|p| 0 <= p < n && spec_prime(p) && n as int %
{ // ------- Model generation 2 starts here ---------
    let mut largest = 1;
    let mut i = 2;
    while i < n
        invariant
            2 <= i <= n,
            1 <= largest <= i,
            spec_prime(largest as int),
            i <= largest ==> spec_prime(i as int),
            largest <= n,
            n %
            forall|p| 0 <= p < i && spec_prime(p) && n as int %
    {
        if is_prime(i) && n %
            largest = i;
        }
        i += 1;
    }
    assert(1 <= largest && largest <= n);
    assert(spec_prime(largest as int));
    assert(n %
    assert(forall|p| 0 <= p < n && spec_prime(p) && n as int %
    largest
}

\end{lstlisting}
\end{figure*}

\begin{figure*}[t!]
    \begin{lstlisting}[language=Verus, caption={Lemma example. \ours{} completes the lemma \texttt{lemma\_step\_subrange}.}, label={lst:lemma-step},basicstyle=\footnotesize\ttfamily]

proof fn lemma_step_subrange(substring: Seq<char>, string: Seq<char>)
    requires
        substring.len() > 0,
        string.len() >= substring.len(),
    ensures
        (substring[0] == string.subrange(0, substring.len() as int)[0] && (substring.skip(1)
            =~= string.skip(1).subrange(0, substring.skip(1).len() as int))) ==> (substring
            =~= string.subrange(0, substring.len() as int)),
    decreases substring.len(),
{ // -------- Model generation starts here -------- 
    let m = substring.len();
    let n = string.len();
    assert(m > 0);
    assert(m <= n);
    let h = string.subrange(0, m as int);
    assert(h.len() == m);
    assert(substring.len() == h.len());
    if substring[0] == h[0] {
        let k = substring.skip(1).len();
        assert(k == h.skip(1).len());
        let g = string.skip(1).subrange(0, k as int);
        assert(g.len() == k);
        assert(k == substring.len() - 1);
        assert(k == h.len() - 1);
        if m > 1 {
            if substring.skip(1) =~= g {
                assert(substring[0] == h[0]);
                assert(substring.len() == g.len() + 1);
                assert(substring.len() == h.len());
                assert(forall|i: nat| 0 <= i && i < m ==> substring[i as int] == h[i as int]) by {
                    assert(substring[0] == h[0]);
                    assert(substring.skip(1) =~= g);
                    assert(forall|i: nat| 1 <= i && i < m ==> substring[i as int] == g[i as int - 1]);
                    assert(substring =~= h);
                }
                assert(substring =~= h);
            } else {
                assert(!(substring =~= h));
            }
        } else {
            assert(substring =~= h);
        }
    } else {
        assert(!(substring =~= h));
    }
}

\end{lstlisting}
\end{figure*}

\begin{algorithm}[ht]
\caption{Iterative Translation and Refinement \pranjal{Need some refinements, but does it look better now? Can put in appendix as well}}
\label{alg:translation}

\KwIn{Source programs $D_{\text{src}}$, initial data $D^{(0)}_{x\to y}, D^{(0)}_{y\to y'}, D^{(0)}_{\text{exploit}}$}
\KwOut{Verified target programs $D_{\text{tgt}}$}

Initialize $i \gets 0$.

\While{not converged}{
  \vspace{0.5em}
  \textbf{(I) Candidate Generation \& Verification:}\\[4pt]
  \ForEach{$x \in D_{\text{src}}$}{
    Generate candidate translations $\{y_j\} \sim G_{\text{explore}}(x; D^{(i)}_{x\to y})$\\
    $C \gets \emptyset$: verified pairs; \quad $S \gets \emptyset$: syntactically correct, unverified candidates\\
    \ForEach{$y_j$}{
      \If{$y_j$ passes verification}{
        $C \gets C \cup \{(x,y_j)\}$
      } \ElseIf{$y_j$ is syntactically correct}{
        $S \gets S \cup \{y_j\}$
      }
    }
  }

  \vspace{0.5em}
  \textbf{(II) Refinement via Treefinement Search:}\\[4pt]
  \ForEach{$y \in S$}{
    Initialize a refinement tree with root node $(y, e(y))$\\
    \While{max iterations not reached}{
      Select node $(y', e(y'))$ by REBASE scoring\\
      Generate refinements $\{y'_k\} \sim G_{\text{refine}}(y', e(y'); D^{(i)}_{y\to y'})$\\
      \ForEach{$y'_k$}{
        \If{$y'_k$ passes verification}{
          $C \gets C \cup \{(x,y'_k)\}$; record trajectory in $C_\tau$\\
          \textbf{break} (stop refining this candidate)
        } \Else {
          Add $(y'_k, e(y'_k))$ as a child node to the refinement tree
        }
      }
    }
  }

  \vspace{0.5em}
  \textbf{(III) Filtering and Data Update:}\\[4pt]
  \ForEach{$(x,y) \in C$}{
    \If{critic rejects $y$ or $f(x,y)=\text{False}$ or exploit model finds $z$ on $s_y$}{
      Discard $y$\\
      \If{exploit model finds $z$}{
        $D^{(i+1)}_{\text{exploit}} \gets D^{(i+1)}_{\text{exploit}} \cup \{(s_y, z)\}$
      }
    }
  }

  Update $D^{(i+1)}_{x \to y} \gets D^{(i)}_{x \to y} \cup C$\\
  Update $D^{(i+1)}_{y \to y'} \gets D^{(i)}_{y \to y'} \cup \{(y, e(y), y')|(x,y') \in C_\tau\}$\\
  $i \gets i + 1$
}

\Return $D_{\text{tgt}} \gets \{y \mid (x,y) \in D^{(i)}_{x \to y}\}$

\end{algorithm}

\end{document}